\documentclass[preprint,12pt,nopreprintline]{elsarticle}
\usepackage[T1]{fontenc}
\usepackage[utf8]{inputenc}
\usepackage{lmodern,textcomp}
\usepackage{amsmath,amssymb,booktabs,array,longtable,graphicx,caption,subcaption,pdflscape}
\usepackage{xurl}
\usepackage[hidelinks]{hyperref}
\biboptions{numbers}
\hypersetup{pdftitle={A Calibration-Referenced Protocol for Measuring Decision Stability under Reference Bank Compression},pdfauthor={Yiwen Chen and Shengming Yu}}
\date{}
\begin{document}
\begin{frontmatter}
\title{A Calibration-Referenced Protocol for Measuring Decision Stability under Reference Bank Compression}

\author{Yiwen Chen\corref{cor1}}
\author{Shengming Yu\corref{cor1}}
\address{Shanghai University of Sport}
\cortext[cor1]{Corresponding authors: Shengming Yu (yushengming@sus.edu.cn) and Yiwen Chen (ywenc236@sus.edu.cn). Primary contact: Shengming Yu.}

\begin{abstract}

Reference banks used for industrial image inspection can be compressed to fit a deployment budget, but the resulting anomaly decisions depend on both the compressed scores and the operating threshold. We present a four-step protocol that compares compression methods at matched deployment sizes, establishes a reference distribution from 200 recalibrations, evaluates disagreement against two calibration benchmarks, and measures changes by distance from the full-bank threshold. With WideResNet-50, budget-matched product quantization (PQ) at the 10\% budget has the lowest decision disagreement among five matched methods, including int8 scalar quantization, on VisA and MVTec AD at both operating levels. Int8 scalar quantization gives lower disagreement than PQ at the other four budgets. On MVTec AD at the lower operating level, PQ meets the calibration median in 34/39 cells and the upper quantile in 39/39 cells. Category-equally weighted image-level AUROC loss at this budget ranges from 0.0003 to 0.0037 across the evaluated configurations. On VisA at the 1\% budget and a 1\% operating level, disagreement reaches 28.12\% (95\% CI 23.25–33.58\%) among near-threshold anomalies and 1.30\% (0.79–2.14\%) across the normal holdout. The 10\% budget-matched PQ setting retains the lowest disagreement among four compared methods with ResNet-18, with clean reference banks, and when thresholds are estimated from normal-directory probes. The protocol connects deployment size to a calibration-referenced decision measure and identifies the evaluation populations most informative for inspection.

\end{abstract}
\begin{keyword}

Reference bank compression; industrial anomaly detection; decision stability; threshold calibration; product quantization; measurement protocol

\end{keyword}
\makeatletter
\global\setbox\keybox=\vbox{\box\keybox}
\makeatother

\end{frontmatter}
\nocite{ref1,ref2,ref3,ref4,ref5,ref6,ref7,ref8,ref9,ref10,ref11,ref12,ref13,ref14,ref15,ref16,ref17,ref18,ref19,ref20,ref21,ref22,ref23,ref24,ref25,ref26,ref27,ref28,ref29,ref30,ref31,ref32,ref33,ref34,ref35,ref36,ref37,ref38,ref39,ref40,ref41,ref42,ref43,ref44,ref45,ref46,ref47,ref48,ref49,ref50,ref51,ref52}
\section{Introduction and related work}
Industrial image inspection converts a score into an action by comparing it with an operating threshold. In embedding-based inspection, distances to a stored reference bank determine the score, while the bank's representation determines deployment storage. Compression changes those distances and can therefore change individual decisions even when image-level ranking is nearly preserved. ROC and precision--recall summaries characterize performance across thresholds \cite{ref1,ref2,ref3}; deployment also requires measuring agreement at the threshold actually used. That agreement depends on threshold estimation: changing the calibration sample can alter decisions even when the full bank remains fixed.

Industrial anomaly-detection benchmarks support comparisons across inspection categories \cite{ref4,ref5,ref6,ref7,ref8}. Existing approaches include pretrained and patch-level features \cite{ref9,ref10,ref11,ref12,ref13,ref14,ref15,ref16,ref17}, feature prediction and distillation \cite{ref18,ref19,ref20,ref21}, flow models \cite{ref22,ref23}, and synthetic defects \cite{ref24}. Noisy references and shared models across classes introduce further questions about which observations define normality \cite{ref25,ref26,ref27,ref28}. Here, the stored feature bank provides a directly measurable storage variable and permits paired comparisons of full-bank and compressed decisions.

Reference selection reduces the number of stored vectors; coverage-oriented methods favor representatives that extend coverage of the feature cloud \cite{ref29,ref30,ref31,ref32}. Such geometric coverage affects image decisions through score aggregation and threshold estimation, motivating evaluation at both ranking and decision levels. Product quantization (PQ) instead partitions vectors into subvectors and stores codes with learned codebooks \cite{ref33,ref34}. Scalar quantization (SQ) trades coordinate precision against retained rows. Equal vector counts therefore do not imply equal storage: comparison requires the bytes of the deployed representation, including its auxiliary structures.

Threshold estimation from normal observations connects the comparison to order statistics \cite{ref35,ref36} and resampling \cite{ref37,ref38}. Distribution-free prediction and conformal methods likewise use calibration samples and finite-sample ranks \cite{ref39,ref40}. Our objective is empirical: measure decision movement under repeated calibration of an unchanged full-bank system, then compare compression disagreement with that distribution on the same evaluation images. Its median and 97.5\% quantile provide typical and upper calibration benchmarks, respectively.

\begin{figure}[tbp]
\centering
\includegraphics[width=\linewidth,height=.70\textheight,keepaspectratio]{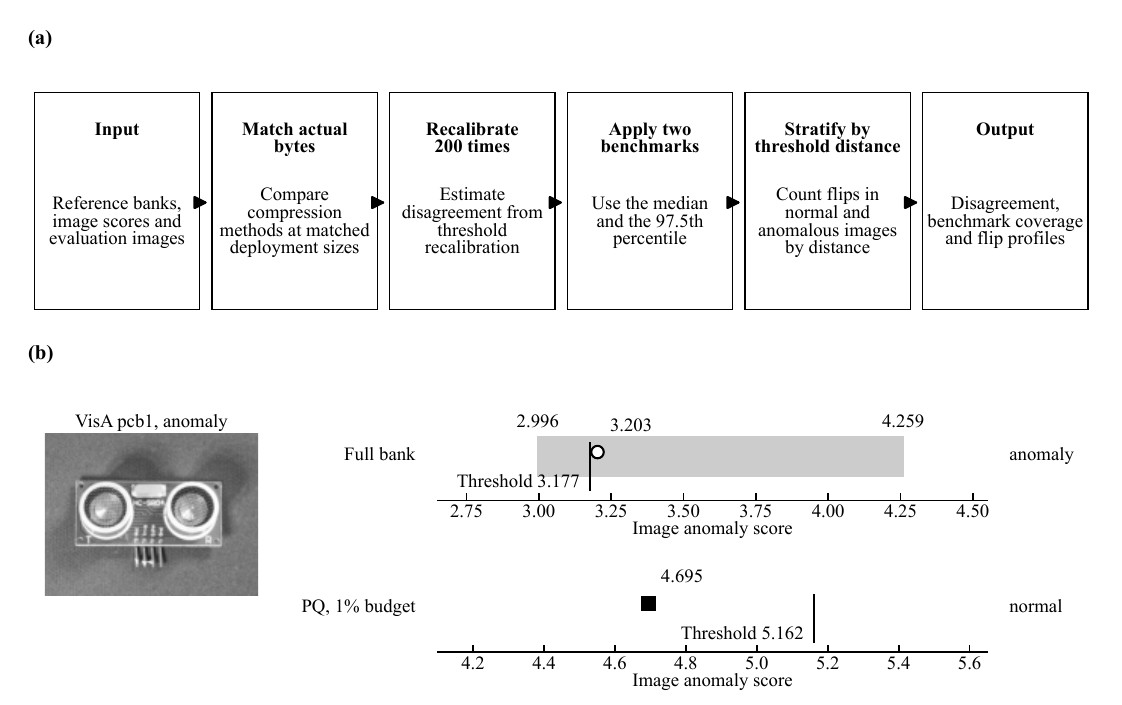}
\caption{Calibration-referenced measurement protocol: match deployment bytes, characterize recalibration variability, compare disagreement with both benchmarks, and stratify decisions by distance from the full-bank threshold. The example is a VisA pcb1 anomaly whose decision changes under 1\% budget-matched PQ at \(\alpha=1\%\). Each row uses its own score scale and threshold; the gray band marks the median to 97.5th percentile of 200 full-bank recalibrated thresholds. The image is resized from VisA (Zou et al., CC BY 4.0).}\label{fig:protocol}
\end{figure}

Measurement studies have examined self-updating reference memory \cite{ref41}, inconsistent labels, appearance shifts and varied product cases \cite{ref42,ref43,ref44,ref45,ref46,ref47,ref48}. The measurement contribution here is a protocol joining resource allocation, threshold variability and decision stability (Fig.~\ref{fig:protocol}). It matches deployment bytes, estimates a reference distribution from 200 recalibrations, evaluates compression disagreement against both benchmarks, and stratifies disagreement by distance from the full-bank threshold. Image-level AUROC and 95\% intervals accompany these measurements.

On VisA and MVTec AD, budget-matched PQ has the lowest disagreement among five matched methods at the 10\% tier, whereas int8 SQ gives lower disagreement than PQ at the other four evaluated tiers. Across evaluated configurations, the 10\% PQ setting loses 0.0003--0.0037 in category-equally weighted image-level AUROC. On VisA with WideResNet-50, a 1\% budget and 1\% operating level yield 28.12\% disagreement among nearest-threshold anomalies versus 1.30\% across the normal holdout. These findings motivate retaining both evaluation populations and checking whether the method ordering persists across backbones and clean banks (Supplementary Tables S4 and S5).

\section{Method}
The protocol operates within a category--seed cell, using one ordered full bank, candidate representations, an original probe set, an independent calibration pool and an evaluation set \(S\). All direct comparisons share the applicable images and role assignments.

\subsection{Deployment budget and representations}
For \(n\) float32 reference vectors of dimension \(d\), the vector payload and nominal cap are
\[
B_0=4nd,\qquad B_{\mathrm{cap}}(q)=\lfloor qB_0\rfloor .
\]
Feasibility uses serialized inference bytes, including the 24-byte header, codes, codebooks and any additional PQ-index bytes. The full bank includes its own header when measured. The fixed feature extractor and transient workspace are outside this reference-bank budget. Realized sizes accompany nominal tiers because row counts and code layouts are discrete (Supplementary Table S6).

Prefix keeps leading rows. Random selection samples without replacement and stores selected rows in bank order. Rebuilt greedy starts with the first row and repeatedly adds the row with greatest squared distance to its nearest selected row, breaking ties by original order. These methods retain as many float32 rows as fit after the header. Budget-matched PQ encodes every row with subspace centroids; asymmetric query distances sum squared distances to the coded centroids \cite{ref33,ref34}. Its layout prioritizes code precision, then payload size, subject to serialized feasibility. Int8 SQ uses the full bank's per-dimension minimum and range, stores these two float32 vectors with signed 8-bit codes for the leading rows that fit, and dequantizes before the same nearest-neighbor search. Fixed-layout lightweight PQ is reported separately at its measured size. Table~\ref{tab:methods} summarizes the comparison; Supplementary Table S7 retains parameters.

\begin{table}[tbp]
\centering
\caption{Reference representations and allocation under a common byte cap. Detailed parameters and realized sizes are in Supplementary Tables S6 and S7.}\label{tab:methods}
\small
\setlength{\tabcolsep}{4pt}
\renewcommand{\arraystretch}{1.15}
\begin{tabular}{@{}>{\raggedright\arraybackslash}p{.22\linewidth}>{\raggedright\arraybackslash}p{.40\linewidth}>{\raggedright\arraybackslash}p{.32\linewidth}@{}}
\toprule
Method & Stored representation & Allocation \\
\midrule
Full bank & All float32 reference vectors & Uncapped reference \\
Prefix & Leading float32 rows & Rows fitting after the header \\
Random & Random float32 subset in bank order & Rows fitting after the header \\
Rebuilt greedy & Coverage-selected float32 subset & Rows fitting after the header \\
Budget-matched PQ & Codes for all rows and subspace codebooks & Feasible layout prioritizes code precision, then payload \\
Int8 SQ & Leading int8 rows and per-dimension minimum/range & Rows fitting with the two float32 vectors \\
Lightweight PQ & Codes for all rows with fixed codebooks & Single measured size; outside the matched tiers \\
\bottomrule
\end{tabular}
\end{table}

\subsection{Thresholds and calibration reference}
Each deployment scores the same original probe and fits its own threshold. For \(m\) ascending probe scores, select rank \(\lceil m(1-\alpha)\rceil\). With \(m=32\), ranks are 32 and 31 at \(\alpha=1\%\) and \(5\%\). A score must strictly exceed its threshold to produce an anomaly decision; ties are normal. The achieved false-positive rate is measured from labeled evaluation images, separately from nominal \(\alpha\).

Hold the full bank fixed and score the independent pool \(P\). The same rank rule gives a whole-pool threshold. Draw 200 sets of 32 images from \(P\), without replacement within each draw; images may recur between draws. Apply each draw's threshold and the whole-pool threshold to the same full-bank evaluation scores. If \(a_{\mathrm{full},r}\) and \(a_{\mathrm{full},P}\) denote these binary decisions, then
\[
D_{\mathrm{cal}}^{(r)}(S)=\frac{1}{|S|}\sum_{x\in S}
I[a_{\mathrm{full},r}(x)\ne a_{\mathrm{full},P}(x)].
\]
Here \(I[\cdot]\) is the indicator. The empirical median and 97.5\% quantile of these 200 values, using type-7 linear interpolation \cite{ref36}, are the calibration benchmarks.

\subsection{Compression disagreement and selection}
Let \(a_b(x)\) be deployment \(b\)'s decision using its original-probe threshold, and \(a_{\mathrm{full}}(x)\) the full bank's decision using its original-probe threshold. Define
\[
D_{\mathrm{comp}}(b;S)=\frac{1}{|S|}\sum_{x\in S}
I[a_b(x)\ne a_{\mathrm{full}}(x)].
\]
Each changed decision counts once, in either direction. Test disagreement uses all test images in the cell, with both labels where available; holdout disagreement uses that cell's normal holdout. Benchmark attainment means \(D_{\mathrm{comp}}\) is no greater than the benchmark for the same population and operating level. Report passing and evaluable cell counts separately from planned coverage (Supplementary Table S10).

Image-level AUROC (I-AUROC) uses labeled test scores \cite{ref2}; its loss is full-bank minus compressed I-AUROC. Macro summaries average seeds within each category and weight categories equally. Direct comparisons use shared available cells. For automatic selection, independently for PQ and prefix, choose the smallest feasible nominal tier meeting the normal-holdout calibration median; use the full bank if all candidates fail. Evaluate the selected deployment on the test set against test calibration benchmarks, as for a prespecified fixed tier (Supplementary Table S11).

\subsection{Uncertainty and paired inference}
Pooled proportions of changed decisions use Wilson 95\% intervals. Category-equally weighted means, attainment fractions and paired differences use 10,000 bootstrap resamples of categories, retaining all seeds of each sampled category; the 2.5\% and 97.5\% percentiles form the interval \cite{ref38}. This preserves clustering across seeds sharing images and defect types. These sampling intervals are distinct from the full-bank recalibration benchmarks. Supplementary Data S1 supplies aggregate intervals.

Paired tests use exact sign flips of category-level differences after seed averaging, with Holm adjustment within each reported family \cite{ref52}. Backbone comparisons against prefix, random and rebuilt greedy form three-test families for each dataset and operating level. Supplementary Tables S2, S4 and S11 retain paired results; Fig.~\ref{fig:paired} displays PQ--SQ comparisons.

\subsection{Threshold-distance strata}
Compute each test image's absolute score distance from the full bank's original-probe threshold. Within each cell, form quintiles separately for truly normal and truly anomalous images, using linearly interpolated boundaries. Assign boundary ties to the nearer-threshold bin. Bin membership is shared across deployments within each cell and label group. Profiles use category-equally weighted bin rates (Fig.~\ref{fig:distance}).

For nearest-bin anomalies versus normal holdouts, pool changed-decision counts and image-occurrence counts over corresponding cells before dividing. The anomaly denominator includes only nearest-bin test anomalies; the holdout denominator includes all normal-holdout images. Each occurrence is counted in its scored cell, so an image can recur across seeds. Whole-test versus holdout comparisons likewise retain separate denominators. Thus the protocol distinguishes disagreement near the decision boundary from agreement on a normal-image proxy.

\section{Experimental setup}
VisA \cite{ref7}, MVTec AD \cite{ref5} and MPDD \cite{ref8} contribute 12, 13 and 6 categories, or 36, 39 and 18 category--seed cells, using seeds 17, 29 and 43. The primary encoder is ImageNet-pretrained WideResNet-50 \cite{ref49,ref50} with PatchCore feature aggregation and bank scoring \cite{ref29}. ResNet-18 \cite{ref51} checks original and clean banks on VisA and MVTec AD. Images are resized to a 256-pixel shorter edge, center-cropped to \(224\times224\) and ImageNet-normalized. Second/third residual-group features form a \(28\times28\) grid of 1,024-dimensional descriptors. Patch scores are nearest-reference squared distances, using asymmetric PQ distances or dequantized SQ vectors. Bilinear interpolation, Gaussian smoothing with 4-pixel standard deviation, and the map maximum yield the image score.

The original PatchCore coreset is the full-bank reference. The original probe estimates deployment thresholds; the independent calibration pool \(P\) supplies whole-pool and resampled thresholds; the normal holdout \(H_{\mathrm n}\) supplies automatic-selection comparisons; and test set \(E\) supplies disagreement, I-AUROC and strata. Pool and holdout are reserved separately from bank fitting, original probes and evaluation; roles and ordered comparison images are fixed across deployments. VisA has 12,965 calibration-pool and 1,152 holdout occurrences; MVTec AD has 4,953 and 1,248. Supplementary Table S8 reports coverage and the available pool inventory, which contains no MPDD records.

Original VisA and MVTec AD banks use 68 fit images per cell: 4 trusted normals and 64 historical candidates. VisA's \texttt{2cls\_highshot} lists include anomaly-directory entries; MVTec AD deliberately injects test-directory defect images into historical lists. Across cells, VisA has 2,448 fit-image occurrences, including 263 anomaly-directory occurrences (10.74\%); MVTec AD has 2,652, including 234 anomaly-directory/injected occurrences (8.82\%). Counts use the union of these labels and count images again when reused across seeds. Removing these occurrences for ResNet-18 clean banks leaves 54--67 fit images per VisA cell and 62 per MVTec AD cell, with zero such entries.

VisA probes contain 135 anomaly-directory occurrences among 1,152; its calibration pool and normal holdout contain none. MVTec AD probes, pool and holdout contain none. A normal-directory-only VisA probe sensitivity check retains 1,017 occurrences, 24--31 per cell (Supplementary Table S1). MPDD banks use normal training images only: 690 fit-image occurrences, comprising 72 trusted normals and 618 historical candidates. Its \texttt{metal\_plate} category has 54 normal training images, constraining independent-pool calibration; at fixed 10\% PQ, attainment distinguishes 15 evaluable from 18 planned cells (Supplementary Table S10).

Primary matched budgets are 0.5\%, 1\%, 5\%, 10\% and 25\% of full-bank vector payload; backbone and clean-bank comparisons use 1\%, 5\% and 10\%. Thresholds are fitted per eligible deployment at \(\alpha=1\%\) and \(5\%\). Ranking is unchanged across operating levels for fixed scores, whereas disagreement depends on the fitted threshold. Supplementary Table S9 provides experimental settings, and S4--S5 provide the backbone and bank comparisons.

\section{Results}
\subsection{Decision disagreement and ranking at matched sizes}
\begin{figure}[tbp]
\centering
\includegraphics[width=\linewidth,height=.70\textheight,keepaspectratio]{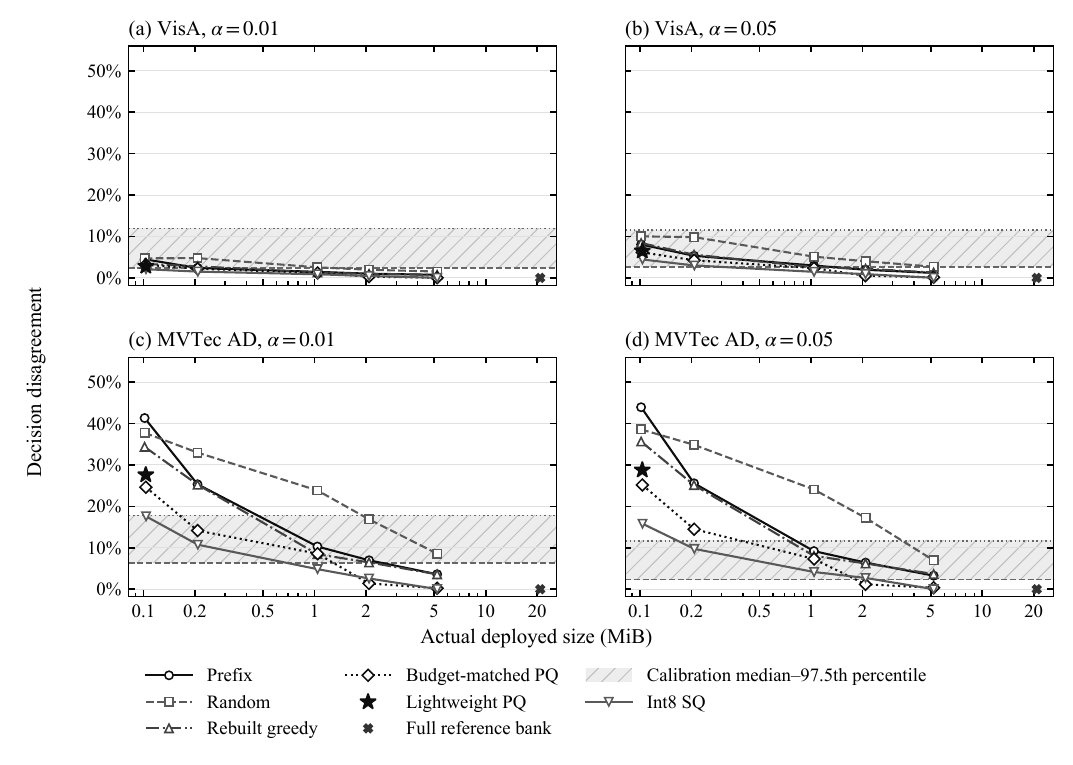}
\caption{Decision disagreement versus deployed size for WideResNet-50. Shading spans the category-equally weighted calibration median and upper-quantile benchmarks. Lightweight PQ is shown at its single realized size; int8 scalar quantization (SQ) uses the same nominal tiers as the matched methods.}\label{fig:disagreement}
\end{figure}

With WideResNet-50, budget-matched PQ had the lowest category-equally weighted decision disagreement among the five matched methods at the 10\% tier on VisA and MVTec AD at both operating levels (Fig.~\ref{fig:disagreement}, Table~\ref{tab:core}). At \(\alpha=1\%\) and \(5\%\), respectively, PQ disagreement was 0.00298 and 0.00529 on VisA, compared with 0.00463 and 0.00921 for int8 SQ. On MVTec AD, the corresponding values were 0.01400 and 0.01193 for PQ, versus 0.02595 and 0.02724 for SQ. Prefix and rebuilt greedy retained more full-bank decisions than random selection at this tier, although PQ was not best at every tier.

\begin{table}[tbp]
\centering
\caption{Core results at the 10\% tier for WideResNet-50 with original banks. Disagreement and I-AUROC loss are category-equally weighted proportions; attainment columns give PQ passing/evaluable cells. Full paired uncertainty is shown in Fig.~\ref{fig:paired} and Supplementary Table S2.}\label{tab:core}
\small
\setlength{\tabcolsep}{3pt}
\renewcommand{\arraystretch}{1.15}
\begin{tabular}{@{}lcccccc@{}}
\toprule
Dataset & $\alpha$ & PQ & SQ & PQ median & PQ upper & PQ I-AUROC \\
 & & $D_{\rm comp}$ & $D_{\rm comp}$ & attainment & attainment & loss \\
\midrule
VisA & 1\% & 0.002984 & 0.004632 & 35/36 & 36/36 & 0.000276 \\
VisA & 5\% & 0.005286 & 0.009206 & 35/36 & 36/36 & 0.000276 \\
MVTec AD & 1\% & 0.01400 & 0.02595 & 34/39 & 39/39 & 0.001513 \\
MVTec AD & 5\% & 0.01193 & 0.02724 & 32/39 & 38/39 & 0.001513 \\
\bottomrule
\end{tabular}
\end{table}

\begin{figure}[tbp]
\centering
\includegraphics[width=\linewidth,height=.70\textheight,keepaspectratio]{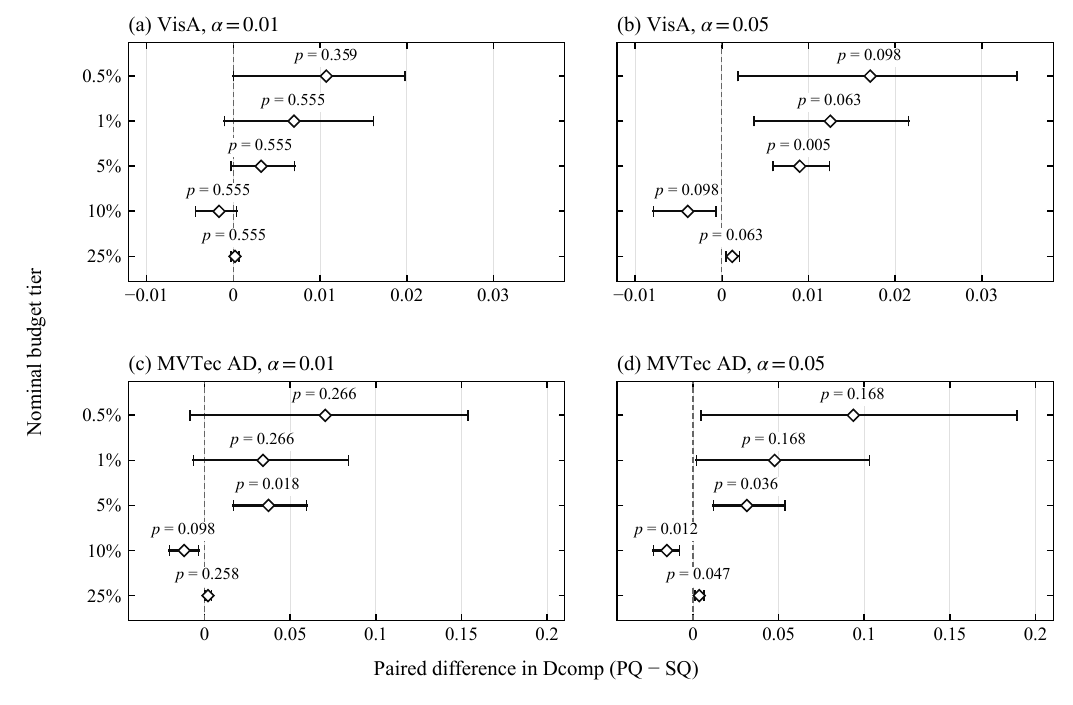}
\caption{Paired differences in decision disagreement between budget-matched PQ and int8 SQ (PQ \(-\) SQ) for WideResNet-50. Horizontal bars are 95\% category-bootstrap intervals; the dashed line marks zero, and labels give Holm-adjusted \(p\)-values.}\label{fig:paired}
\end{figure}

The paired PQ-minus-SQ differences at 10\% were \(-0.0016\) (95\% CI \(-0.0043\) to 0.0004) and \(-0.0039\) (\(-0.0079\) to \(-0.0007\)) on VisA, and \(-0.0120\) (\(-0.0205\) to \(-0.0035\)) and \(-0.0153\) (\(-0.0231\) to \(-0.0080\)) on MVTec AD, in ascending operating-level order. Holm-adjusted \(p\)-values were 0.555, 0.098, 0.098 and 0.012; only the last contrast had adjusted \(p<0.05\). Thus the observed minimum does not imply a significant advantage in every comparison. At the other four tiers, SQ had lower disagreement in all 16 VisA and MVTec AD comparisons. At 25\%, its disagreement was at most 0.00046. Figure~\ref{fig:paired} and Supplementary Table S2 retain the intervals and adjusted tests across tiers.

At 10\%, PQ occupied 2,180,360 bytes on both datasets. Its I-AUROC was 0.9111 on VisA and 0.9759 on MVTec AD, against 0.9114 and 0.9774 for the full banks. SQ occupied 2,183,192 bytes and achieved 0.9117 and 0.9769, respectively: similar ranking accompanied different decisions at the operating thresholds. Across the evaluated backbone, bank and dataset configurations, PQ I-AUROC loss at 10\% ranged from 0.000276 to 0.003715; the upper limits of the seven 95\% category-bootstrap intervals ranged from 0.0018 to 0.0070 (Supplementary Data S1). Each loss uses the full bank from the same configuration. Supplementary Fig.~S1 and Tables S4--S6 provide the detailed ranking comparisons, including the separate lightweight-PQ operating point.

\subsection{Calibration attainment and budget selection}
\begin{figure}[tbp]
\centering
\includegraphics[width=\linewidth,height=.70\textheight,keepaspectratio]{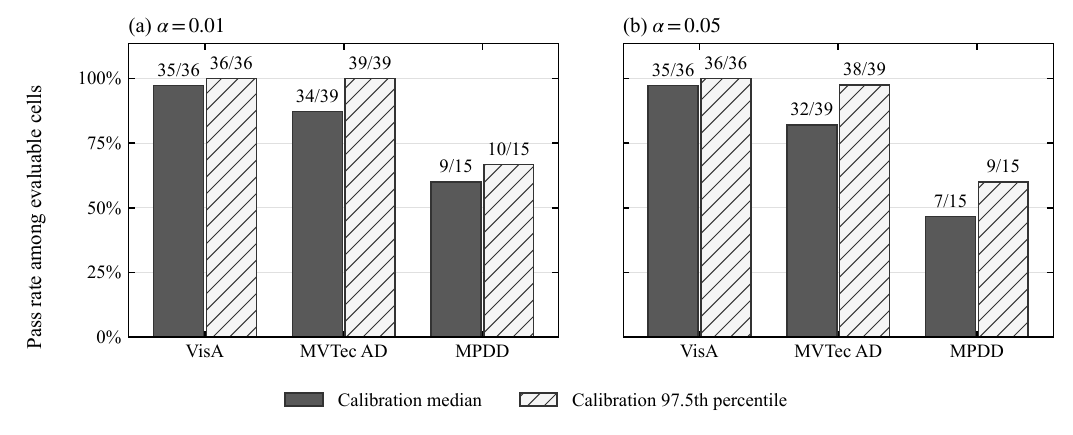}
\caption{Cell-level calibration attainment for fixed 10\% budget-matched PQ. The median and upper-quantile benchmarks are evaluated separately for each category--seed cell.}\label{fig:attainment}
\end{figure}

The calibration median describes typical decision movement under recalibration, whereas the 97.5\% quantile provides a broader empirical benchmark. At fixed 10\% PQ, VisA met the median in 35/36 cells and the upper quantile in 36/36 at both operating levels. MVTec AD met them in 34/39 and 39/39 cells at \(\alpha=1\%\), and 32/39 and 38/39 at \(5\%\). MPDD attained the median in 9/15 and 7/15 evaluable cells, and the upper quantile in 10/15 and 9/15, respectively (Fig.~\ref{fig:attainment}). PQ had higher median attainment than 10\% SQ on VisA and MVTec AD; at 25\%, SQ met both benchmarks in every cell of these datasets. Supplementary Table S10 preserves the planned and evaluable denominators.

Automatic PQ selection chooses the smallest available budget satisfying the normal-holdout calibration median, falling back to the full bank if necessary. Compared with fixed 10\% PQ, its paired test-set disagreement was higher by 0.0179 and 0.0269 on VisA at the two operating levels, with Holm-adjusted \(p=0.00195\) for each. MVTec AD differences were 0.0892 and 0.0914, with adjusted \(p=0.000977\) and 0.00391. MPDD differences were 0.0517 and 0.0659, but both intervals included zero and adjusted \(p\)-values were 0.750 and 0.563. These comparisons concern test decisions after selection from normal-image evidence; they do not establish that every selected budget is inferior. Supplementary Table S11 also retains comparisons against fixed 1\% PQ.

\subsection{Disagreement near the threshold}
Normal-holdout agreement did not characterize the behavior of near-threshold anomalies. At fixed 1\% PQ and \(\alpha=1\%\) on VisA, WideResNet-50 produced 28.12\% disagreement among nearest-bin anomalies (81/288; 95\% CI 23.25--33.58\%), versus 1.30\% across the normal holdout (15/1,152; 0.79--2.14\%). With automatically selected PQ budgets, the corresponding pooled anomaly rates were 31.60\% on VisA (91/288), 26.91\% on MVTec AD (176/654) and 23.48\% on MPDD (31/132), versus normal-holdout rates of 0.17\% (2/1,152), 0.24\% (3/1,248) and 0\% (0/480). Denominators count image occurrences across category--seed combinations; an image can recur across seeds.

The nearest threshold-distance bin had the highest category-equally weighted disagreement in all 36 plotted curves, although only 28 were nonincreasing with distance (Fig.~\ref{fig:distance}). These category-weighted profiles differ from the pooled rates above. Each deployment uses its own fitted threshold, so changed decisions reflect both score and threshold displacement. Under automatic PQ selection on MVTec AD at \(\alpha=1\%\), normal-holdout and whole-test disagreement had Spearman correlation \(-0.090\). For \texttt{metal\_nut}, seed 29, normal-holdout agreement was complete while 71/109 test decisions changed (Supplementary Fig.~S3). The two populations therefore supply complementary evidence for reviewing a compressed deployment.

\begin{figure}[tbp]
\centering
\includegraphics[width=\linewidth,height=.70\textheight,keepaspectratio]{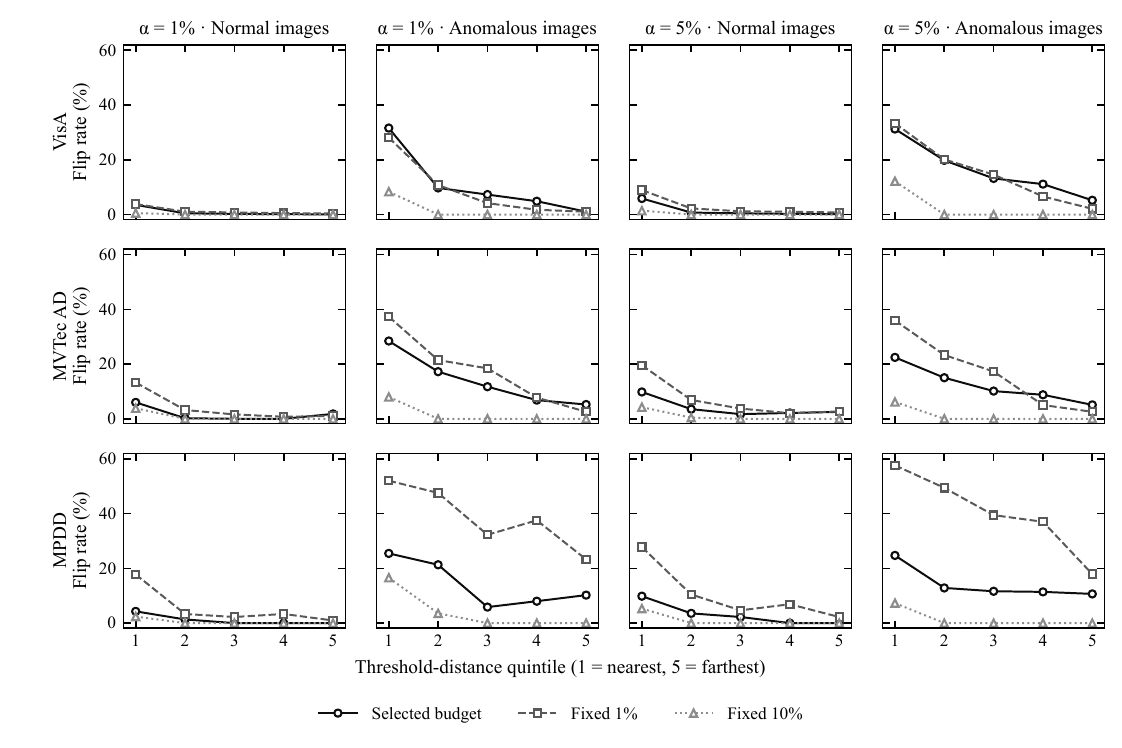}
\caption{Decision disagreement stratified by distance from the full-bank threshold, separately for truly normal and truly anomalous test images. The category-equally weighted profiles use bins defined within each category--seed cell. The whole-test versus normal-holdout comparison is provided in Supplementary Fig.~S3.}\label{fig:distance}
\end{figure}

\section{Validation}
The ResNet-18 original-bank comparison covered VisA and MVTec AD at 1\%, 5\% and 10\% budgets. Three checks assessed the lowest disagreement at 10\%, joint calibration attainment at 10\%, and higher nearest-bin anomaly than normal-holdout disagreement at 1\%. Joint attainment required at least 80\% of evaluable cells to meet the median and 90\% to meet the upper quantile. All three checks held in all four dataset-by-operating-level combinations. The PQ minimum in these validation comparisons is among four methods: PQ, prefix, random and rebuilt greedy (Supplementary Table S4 and Fig.~S4).

At 10\%, ResNet-18 PQ disagreement was 0.00255 and 0.00623 on VisA, and 0.01477 and 0.02375 on MVTec AD, at the lower and higher operating levels. VisA met the median in 35/36 cells at both levels; MVTec AD did so in 34/39 and 33/39. All cells met the upper quantile. At fixed 1\% and the lower operating level, VisA nearest-bin anomaly disagreement was 26.39\% (76/288), against 0.87\% (10/1,152) on the normal holdout. Paired exact sign-flip tests against the three subset methods used separate Holm families for each dataset and operating level; adjusted \(p\)-values ranged from 0.000732 to 0.00586 (Supplementary Table S4) \cite{ref52}.

The ResNet-18 clean-bank comparison also met all three checks in all four combinations, using its own clean full-bank scores and calibration references. PQ disagreement at 10\% was 0.00229 and 0.00555 on VisA, and 0.01526 and 0.01628 on MVTec AD. Median attainment was 35/36 and 36/36 on VisA, and 32/39 and 36/39 on MVTec AD; all cells met the upper quantile (Supplementary Table S5).

Restricting VisA threshold probes to normal directories retained 1,017 image occurrences across 36 cells, with 24--31 per cell. At both operating levels, PQ remained the lowest-disagreement method among the four compared methods for the WideResNet-50 original bank and both ResNet-18 banks. Median attainment ranged from 28/36 to 33/36, upper-quantile attainment was always 36/36, and PQ disagreement ranged from 0.00954 to 0.01659 (Supplementary Table S1).

\section{Discussion}
Deployment size, score ranking and threshold decisions describe different properties of a compressed bank. The calibration reference makes disagreement interpretable against decision movement caused by refitting the full-bank threshold. Meeting the median and meeting the upper quantile are distinct readings, so both should accompany cell-level counts. A change in calibration practice requires a corresponding reference distribution.

The storage-dependent ordering is consistent with the representations. At 10\%, SQ retains about 40\% of bank rows, whereas PQ encodes every row; at 25\%, SQ retains nearly all rows. The fixed 10\% PQ setting is therefore a starting point for assessment within the evaluated configurations. A deployment review should fit the compressed-bank threshold, record actual bytes and I-AUROC, and compare paired decisions on representative images. Normal holdouts characterize nominal parts; labeled anomalies close to the full-bank threshold reveal changes among borderline defects. Reporting both rates and denominators, together with category-level results, supports product-specific review. Supplementary Fig.~S2 gives the category-by-budget breakdown.

\subsection{Limitations}
MPDD \texttt{metal\_plate} has only 54 normal training images, and the independent-pool requirement restricts its calibration evaluation. A separate verified normal production pool would permit the comparison with bank and holdout roles separated. Backbone and clean-bank checks cover WideResNet-50 and ResNet-18 on VisA and MVTec AD; the normal-directory probe check is confined to VisA. Other feature representations or distances require matching realized bytes and re-establishing their calibration references. Category-equally weighted summaries describe the evaluated category set; product-level decisions require the corresponding category and seed results.

\section{Conclusion}
The protocol connects matched deployment size to calibration-referenced decision disagreement and its location relative to the full-bank threshold. In the primary VisA and MVTec AD comparison, 10\% budget-matched PQ had the lowest observed disagreement, while SQ performed better at the other evaluated tiers. Backbone, clean-bank and probe-source comparisons supported the stated 10\% finding within their evaluated method sets. Near-threshold anomalies changed decisions much more often than normal holdouts, showing why ranking, calibration attainment and both evaluation populations should be reported together when assessing compression.

\section*{Funding}
This work was supported by first-year funding awarded to Shengming Yu under the 2025 Shanghai Higher Education Young Teachers Cultivation Funding Program (funding reference A1-0202-26-0024-13).

\section*{Acknowledgments}
The authors thank the contributors to the VisA, MVTec AD, and MPDD datasets for making these resources publicly available.

\section*{CRediT authorship contribution statement}
Yiwen Chen: Conceptualization, Methodology, Software, Investigation, Formal analysis, Visualization, Writing -- original draft, Writing -- review \& editing.

Shengming Yu: Methodology, Supervision, Funding acquisition.

\section*{Declaration of competing interest}
The authors declare that they have no known competing financial interests or personal relationships that could have appeared to influence the work reported in this paper.

\section*{Data availability}
This study used the publicly available VisA \cite{ref7}, MVTec AD \cite{ref5}, and MPDD \cite{ref8} datasets. Supplementary Data S1, containing interval summaries, is provided as an ancillary workbook with this preprint. The study's code and per-image scores are not included in this preprint deposit.

\section*{Declaration of generative AI and AI-assisted technologies}
During the preparation of this work, the authors used OpenAI Codex to assist with language editing, manuscript condensation, and organization of supplementary and submission materials. The authors take full responsibility for the content of this article.

\bibliographystyle{elsarticle-num}
\bibliography{references}

\clearpage
\renewcommand{\thetable}{S\arabic{table}}
\renewcommand{\thefigure}{S\arabic{figure}}
\renewcommand{\theHtable}{supp.\arabic{table}}
\renewcommand{\theHfigure}{supp.\arabic{figure}}
\setcounter{table}{0}
\setcounter{figure}{0}
\section*{Supplementary material}

This supplement provides Tables S1--S11 and Figs. S1--S4. The supplementary workbook, Supplementary Data S1, reports the measurement intervals and is provided as an ancillary file with this preprint. The study's code and per-image scores are not included in this preprint deposit.

\clearpage

\begin{landscape}

\section{Probe-source sensitivity}

The comparison covers the WideResNet-50 original bank and the ResNet-18 original and clean banks at the 10\% budget. Threshold estimation with normal-directory probes retained 1,017 of 1,152 image occurrences across 36 category\textendash{}seed cells, with 24\textendash{}31 retained per cell. The 135 excluded occurrences came from anomaly directories. Dcomp values are proportions displayed to six decimal places. J2 columns report the number of cells meeting each calibration benchmark.

\begingroup

\footnotesize

\setlength{\tabcolsep}{3pt}

\renewcommand{\arraystretch}{1.15}

\setcounter{table}{0}

\begin{longtable}{>{\raggedright\arraybackslash}p{\dimexpr 0.170000\linewidth-9.180000pt\relax}>{\raggedright\arraybackslash}p{\dimexpr 0.050000\linewidth-2.700000pt\relax}>{\raggedright\arraybackslash}p{\dimexpr 0.170000\linewidth-9.180000pt\relax}>{\raggedright\arraybackslash}p{\dimexpr 0.090000\linewidth-4.860000pt\relax}>{\raggedright\arraybackslash}p{\dimexpr 0.090000\linewidth-4.860000pt\relax}>{\raggedright\arraybackslash}p{\dimexpr 0.110000\linewidth-5.940000pt\relax}>{\raggedright\arraybackslash}p{\dimexpr 0.130000\linewidth-7.020000pt\relax}>{\raggedright\arraybackslash}p{\dimexpr 0.080000\linewidth-4.320000pt\relax}>{\raggedright\arraybackslash}p{\dimexpr 0.110000\linewidth-5.940000pt\relax}}

\caption{VisA sensitivity to probe-image source}\label{tab:s1}\\

\toprule

Backbone and bank & \ensuremath{\alpha} & Probe source & Prefix Dcomp & Random Dcomp & Rebuilt greedy Dcomp & Budget-matched PQ Dcomp & J2 median & J2 97.5th percentile \\

\midrule

\endfirsthead

\multicolumn{9}{l}{\textit{Table S1 continued}}\\

\toprule

Backbone and bank & \ensuremath{\alpha} & Probe source & Prefix Dcomp & Random Dcomp & Rebuilt greedy Dcomp & Budget-matched PQ Dcomp & J2 median & J2 97.5th percentile \\

\midrule

\endhead

\bottomrule

\endfoot

WideResNet-50 original & 0.01 & All recorded probe images & 0.010220 & 0.019702 & 0.009629 & 0.002984 & 35/36 & 36/36 \\
WideResNet-50 original & 0.01 & Normal directories & 0.034941 & 0.065929 & 0.042914 & 0.010088 & 33/36 & 36/36 \\
WideResNet-50 original & 0.05 & All recorded probe images & 0.019600 & 0.040281 & 0.022073 & 0.005286 & 35/36 & 36/36 \\
WideResNet-50 original & 0.05 & Normal directories & 0.053127 & 0.095792 & 0.060188 & 0.015700 & 28/36 & 36/36 \\
ResNet-18 original & 0.01 & All recorded probe images & 0.009114 & 0.020931 & 0.008909 & 0.002550 & 35/36 & 36/36 \\
ResNet-18 original & 0.01 & Normal directories & 0.041193 & 0.064180 & 0.036387 & 0.009535 & 33/36 & 36/36 \\
ResNet-18 original & 0.05 & All recorded probe images & 0.020801 & 0.044239 & 0.019923 & 0.006228 & 35/36 & 36/36 \\
ResNet-18 original & 0.05 & Normal directories & 0.059521 & 0.100652 & 0.064620 & 0.014936 & 31/36 & 36/36 \\
ResNet-18 clean & 0.01 & All recorded probe images & 0.006733 & 0.017033 & 0.008164 & 0.002292 & 35/36 & 36/36 \\
ResNet-18 clean & 0.01 & Normal directories & 0.036625 & 0.066954 & 0.033243 & 0.009690 & 31/36 & 36/36 \\
ResNet-18 clean & 0.05 & All recorded probe images & 0.018657 & 0.040122 & 0.015495 & 0.005549 & 36/36 & 36/36 \\
ResNet-18 clean & 0.05 & Normal directories & 0.057318 & 0.111017 & 0.054412 & 0.016588 & 30/36 & 36/36 \\

\end{longtable}

\endgroup

Budget-matched PQ gives the lowest Dcomp at both operating levels for all three configurations after restricting the probes to normal directories. For the WideResNet-50 original bank, the median benchmark is met in 33/36 cells at \ensuremath{\alpha} = 1\% and 28/36 at \ensuremath{\alpha} = 5\%. The latter is below the 80\% median criterion for joint J2 attainment. The 97.5th-percentile benchmark is met in 36/36 cells at both levels.

The probe restriction changes both the source composition and the number of images used to estimate each threshold. The two calibration benchmarks for each bank use its existing independent calibration pool and recorded resampling indices. The accompanying CSV supplies the displayed results at full precision and the per-cell CSV records the inputs to the reported counts.

\end{landscape}

\clearpage

\begin{landscape}

\section{Product and scalar quantization comparisons}

\begingroup

\footnotesize

\setlength{\tabcolsep}{3pt}

\renewcommand{\arraystretch}{1.15}

\setcounter{table}{1}

\begin{longtable}{>{\raggedright\arraybackslash}p{\dimexpr 0.105000\linewidth-5.670000pt\relax}>{\raggedright\arraybackslash}p{\dimexpr 0.060000\linewidth-3.240000pt\relax}>{\raggedright\arraybackslash}p{\dimexpr 0.070000\linewidth-3.780000pt\relax}>{\raggedright\arraybackslash}p{\dimexpr 0.100000\linewidth-5.400000pt\relax}>{\raggedright\arraybackslash}p{\dimexpr 0.100000\linewidth-5.400000pt\relax}>{\raggedright\arraybackslash}p{\dimexpr 0.105000\linewidth-5.670000pt\relax}>{\raggedright\arraybackslash}p{\dimexpr 0.255000\linewidth-13.770000pt\relax}>{\raggedright\arraybackslash}p{\dimexpr 0.105000\linewidth-5.670000pt\relax}>{\raggedright\arraybackslash}p{\dimexpr 0.100000\linewidth-5.400000pt\relax}}

\caption{Paired comparisons of fixed budget-matched PQ and fixed int8 SQ (PQ \ensuremath{-} SQ in Dcomp) for all tiers, datasets and operating levels.}\label{tab:s2}\\

\toprule

Dataset & \ensuremath{\alpha} & Tier & PQ Dcomp & SQ Dcomp & Mean difference & 95\% CI & Holm p & Categories \\

\midrule

\endfirsthead

\multicolumn{9}{l}{\textit{Table S2 continued}}\\

\toprule

Dataset & \ensuremath{\alpha} & Tier & PQ Dcomp & SQ Dcomp & Mean difference & 95\% CI & Holm p & Categories \\

\midrule

\endhead

\bottomrule

\endfoot

MPDD & 0.01 & 0.5\% & 0.1755 & 0.1451 & 0.03041 & [\ensuremath{-}0.008439, 0.06926] & 1.00 & 2 \\
MPDD & 0.01 & 1\% & 0.2292 & 0.1605 & 0.06864 & [0.04091, 0.1027] & 0.313 & 5 \\
MPDD & 0.01 & 5\% & 0.07760 & 0.05831 & 0.01930 & [\ensuremath{-}0.004547, 0.04837] & 1.00 & 6 \\
MPDD & 0.01 & 10\% & 0.02755 & 0.03935 & \ensuremath{-}0.01180 & [\ensuremath{-}0.03396, 0.002786] & 1.00 & 6 \\
MPDD & 0.01 & 25\% & 0.006845 & 0 & 0.006845 & [0.0007032, 0.01455] & 1.00 & 6 \\
MPDD & 0.05 & 0.5\% & 0.2837 & 0.2356 & 0.04811 & [\ensuremath{-}0.03797, 0.1342] & 1.00 & 2 \\
MPDD & 0.05 & 1\% & 0.2473 & 0.1671 & 0.08017 & [\ensuremath{-}0.006674, 0.1585] & 0.750 & 5 \\
MPDD & 0.05 & 5\% & 0.05881 & 0.04719 & 0.01162 & [\ensuremath{-}0.01907, 0.05398] & 1.00 & 6 \\
MPDD & 0.05 & 10\% & 0.01543 & 0.03617 & \ensuremath{-}0.02074 & [\ensuremath{-}0.04146, \ensuremath{-}0.003297] & 0.625 & 6 \\
MPDD & 0.05 & 25\% & 0.004765 & 0 & 0.004765 & [0.001263, 0.008654] & 0.750 & 6 \\
MVTec AD & 0.01 & 0.5\% & 0.2463 & 0.1759 & 0.07044 & [\ensuremath{-}0.008534, 0.1538] & 0.266 & 13 \\
MVTec AD & 0.01 & 1\% & 0.1416 & 0.1075 & 0.03410 & [\ensuremath{-}0.006503, 0.08398] & 0.266 & 13 \\
MVTec AD & 0.01 & 5\% & 0.08571 & 0.04842 & 0.03729 & [0.01674, 0.05957] & 0.0183 & 13 \\
MVTec AD & 0.01 & 10\% & 0.01400 & 0.02595 & \ensuremath{-}0.01196 & [\ensuremath{-}0.02048, \ensuremath{-}0.003545] & 0.0977 & 13 \\
MVTec AD & 0.01 & 25\% & 0.002464 & 0.0004508 & 0.002013 & [0.0001204, 0.004078] & 0.258 & 13 \\
MVTec AD & 0.05 & 0.5\% & 0.2518 & 0.1581 & 0.09361 & [0.004675, 0.1891] & 0.168 & 13 \\
MVTec AD & 0.05 & 1\% & 0.1450 & 0.09738 & 0.04764 & [0.001973, 0.1030] & 0.168 & 13 \\
MVTec AD & 0.05 & 5\% & 0.07305 & 0.04162 & 0.03143 & [0.01187, 0.05357] & 0.0361 & 13 \\
MVTec AD & 0.05 & 10\% & 0.01193 & 0.02724 & \ensuremath{-}0.01531 & [\ensuremath{-}0.02311, \ensuremath{-}0.007973] & 0.0122 & 13 \\
MVTec AD & 0.05 & 25\% & 0.003781 & 0.0001665 & 0.003614 & [0.001263, 0.006525] & 0.0469 & 13 \\
VisA & 0.01 & 0.5\% & 0.03176 & 0.02103 & 0.01073 & [0.00001590, 0.01982] & 0.359 & 12 \\
VisA & 0.01 & 1\% & 0.02218 & 0.01518 & 0.006997 & [\ensuremath{-}0.001039, 0.01616] & 0.555 & 12 \\
VisA & 0.01 & 5\% & 0.01196 & 0.008754 & 0.003207 & [\ensuremath{-}0.0002808, 0.007079] & 0.555 & 12 \\
VisA & 0.01 & 10\% & 0.002984 & 0.004632 & \ensuremath{-}0.001648 & [\ensuremath{-}0.004342, 0.0003626] & 0.555 & 12 \\
VisA & 0.01 & 25\% & 0.0004938 & 0.0003046 & 0.0001892 & [\ensuremath{-}0.0002838, 0.0006617] & 0.555 & 12 \\
VisA & 0.05 & 0.5\% & 0.06157 & 0.04445 & 0.01712 & [0.001847, 0.03404] & 0.0977 & 12 \\
VisA & 0.05 & 1\% & 0.04256 & 0.03003 & 0.01253 & [0.003731, 0.02155] & 0.0625 & 12 \\
VisA & 0.05 & 5\% & 0.02364 & 0.01465 & 0.008993 & [0.005886, 0.01242] & 0.00488 & 12 \\
VisA & 0.05 & 10\% & 0.005286 & 0.009206 & \ensuremath{-}0.003920 & [\ensuremath{-}0.007878, \ensuremath{-}0.0006536] & 0.0977 & 12 \\
VisA & 0.05 & 25\% & 0.001389 & 0.0001894 & 0.001200 & [0.0004920, 0.002038] & 0.0625 & 12 \\

\end{longtable}

\endgroup

\end{landscape}

\clearpage

\begin{landscape}

\section{Scalar-quantization calibration attainment}

\begingroup

\footnotesize

\setlength{\tabcolsep}{3pt}

\renewcommand{\arraystretch}{1.15}

\setcounter{table}{2}

\begin{longtable}{>{\raggedright\arraybackslash}p{\dimexpr 0.085000\linewidth-5.610000pt\relax}>{\raggedright\arraybackslash}p{\dimexpr 0.055000\linewidth-3.630000pt\relax}>{\raggedright\arraybackslash}p{\dimexpr 0.110000\linewidth-7.260000pt\relax}>{\raggedright\arraybackslash}p{\dimexpr 0.150000\linewidth-9.900000pt\relax}>{\raggedright\arraybackslash}p{\dimexpr 0.050000\linewidth-3.300000pt\relax}>{\raggedright\arraybackslash}p{\dimexpr 0.075000\linewidth-4.950000pt\relax}>{\raggedright\arraybackslash}p{\dimexpr 0.075000\linewidth-4.950000pt\relax}>{\raggedright\arraybackslash}p{\dimexpr 0.045000\linewidth-2.970000pt\relax}>{\raggedright\arraybackslash}p{\dimexpr 0.085000\linewidth-5.610000pt\relax}>{\raggedright\arraybackslash}p{\dimexpr 0.135000\linewidth-8.910000pt\relax}>{\raggedright\arraybackslash}p{\dimexpr 0.135000\linewidth-8.910000pt\relax}}

\caption{Calibration benchmark attainment for fixed int8 SQ at all tiers.}\label{tab:s3}\\

\toprule

Dataset & \ensuremath{\alpha} & Benchmark & Configuration & Pass & Eligible & Planned & NA & Fallback & Eligible rate & Planned rate \\

\midrule

\endfirsthead

\multicolumn{11}{l}{\textit{Table S3 continued}}\\

\toprule

Dataset & \ensuremath{\alpha} & Benchmark & Configuration & Pass & Eligible & Planned & NA & Fallback & Eligible rate & Planned rate \\

\midrule

\endhead

\bottomrule

\endfoot

MPDD & 0.01 & Median & Fixed SQ 0.5\% & 2 & 15 & 18 & 3 & 0 & 13.33\% & 11.11\% \\
MPDD & 0.01 & Upper quantile & Fixed SQ 0.5\% & 6 & 15 & 18 & 3 & 0 & 40.00\% & 33.33\% \\
MPDD & 0.01 & Median & Fixed SQ 1\% & 3 & 15 & 18 & 3 & 0 & 20.00\% & 16.67\% \\
MPDD & 0.01 & Upper quantile & Fixed SQ 1\% & 7 & 15 & 18 & 3 & 0 & 46.67\% & 38.89\% \\
MPDD & 0.01 & Median & Fixed SQ 5\% & 5 & 15 & 18 & 3 & 0 & 33.33\% & 27.78\% \\
MPDD & 0.01 & Upper quantile & Fixed SQ 5\% & 9 & 15 & 18 & 3 & 0 & 60.00\% & 50.00\% \\
MPDD & 0.01 & Median & Fixed SQ 10\% & 8 & 15 & 18 & 3 & 0 & 53.33\% & 44.44\% \\
MPDD & 0.01 & Upper quantile & Fixed SQ 10\% & 11 & 15 & 18 & 3 & 0 & 73.33\% & 61.11\% \\
MPDD & 0.01 & Median & Fixed SQ 25\% & 15 & 15 & 18 & 3 & 0 & 100.00\% & 83.33\% \\
MPDD & 0.01 & Upper quantile & Fixed SQ 25\% & 15 & 15 & 18 & 3 & 0 & 100.00\% & 83.33\% \\
MPDD & 0.05 & Median & Fixed SQ 0.5\% & 1 & 15 & 18 & 3 & 0 & 6.67\% & 5.56\% \\
MPDD & 0.05 & Upper quantile & Fixed SQ 0.5\% & 3 & 15 & 18 & 3 & 0 & 20.00\% & 16.67\% \\
MPDD & 0.05 & Median & Fixed SQ 1\% & 1 & 15 & 18 & 3 & 0 & 6.67\% & 5.56\% \\
MPDD & 0.05 & Upper quantile & Fixed SQ 1\% & 3 & 15 & 18 & 3 & 0 & 20.00\% & 16.67\% \\
MPDD & 0.05 & Median & Fixed SQ 5\% & 3 & 15 & 18 & 3 & 0 & 20.00\% & 16.67\% \\
MPDD & 0.05 & Upper quantile & Fixed SQ 5\% & 8 & 15 & 18 & 3 & 0 & 53.33\% & 44.44\% \\
MPDD & 0.05 & Median & Fixed SQ 10\% & 4 & 15 & 18 & 3 & 0 & 26.67\% & 22.22\% \\
MPDD & 0.05 & Upper quantile & Fixed SQ 10\% & 8 & 15 & 18 & 3 & 0 & 53.33\% & 44.44\% \\
MPDD & 0.05 & Median & Fixed SQ 25\% & 15 & 15 & 18 & 3 & 0 & 100.00\% & 83.33\% \\
MPDD & 0.05 & Upper quantile & Fixed SQ 25\% & 15 & 15 & 18 & 3 & 0 & 100.00\% & 83.33\% \\
MVTec AD & 0.01 & Median & Fixed SQ 0.5\% & 9 & 39 & 39 & 0 & 0 & 23.08\% & 23.08\% \\
MVTec AD & 0.01 & Upper quantile & Fixed SQ 0.5\% & 22 & 39 & 39 & 0 & 0 & 56.41\% & 56.41\% \\
MVTec AD & 0.01 & Median & Fixed SQ 1\% & 13 & 39 & 39 & 0 & 0 & 33.33\% & 33.33\% \\
MVTec AD & 0.01 & Upper quantile & Fixed SQ 1\% & 28 & 39 & 39 & 0 & 0 & 71.79\% & 71.79\% \\
MVTec AD & 0.01 & Median & Fixed SQ 5\% & 28 & 39 & 39 & 0 & 0 & 71.79\% & 71.79\% \\
MVTec AD & 0.01 & Upper quantile & Fixed SQ 5\% & 37 & 39 & 39 & 0 & 0 & 94.87\% & 94.87\% \\
MVTec AD & 0.01 & Median & Fixed SQ 10\% & 32 & 39 & 39 & 0 & 0 & 82.05\% & 82.05\% \\
MVTec AD & 0.01 & Upper quantile & Fixed SQ 10\% & 38 & 39 & 39 & 0 & 0 & 97.44\% & 97.44\% \\
MVTec AD & 0.01 & Median & Fixed SQ 25\% & 39 & 39 & 39 & 0 & 0 & 100.00\% & 100.00\% \\
MVTec AD & 0.01 & Upper quantile & Fixed SQ 25\% & 39 & 39 & 39 & 0 & 0 & 100.00\% & 100.00\% \\
MVTec AD & 0.05 & Median & Fixed SQ 0.5\% & 4 & 39 & 39 & 0 & 0 & 10.26\% & 10.26\% \\
MVTec AD & 0.05 & Upper quantile & Fixed SQ 0.5\% & 18 & 39 & 39 & 0 & 0 & 46.15\% & 46.15\% \\
MVTec AD & 0.05 & Median & Fixed SQ 1\% & 5 & 39 & 39 & 0 & 0 & 12.82\% & 12.82\% \\
MVTec AD & 0.05 & Upper quantile & Fixed SQ 1\% & 23 & 39 & 39 & 0 & 0 & 58.97\% & 58.97\% \\
MVTec AD & 0.05 & Median & Fixed SQ 5\% & 13 & 39 & 39 & 0 & 0 & 33.33\% & 33.33\% \\
MVTec AD & 0.05 & Upper quantile & Fixed SQ 5\% & 36 & 39 & 39 & 0 & 0 & 92.31\% & 92.31\% \\
MVTec AD & 0.05 & Median & Fixed SQ 10\% & 21 & 39 & 39 & 0 & 0 & 53.85\% & 53.85\% \\
MVTec AD & 0.05 & Upper quantile & Fixed SQ 10\% & 39 & 39 & 39 & 0 & 0 & 100.00\% & 100.00\% \\
MVTec AD & 0.05 & Median & Fixed SQ 25\% & 39 & 39 & 39 & 0 & 0 & 100.00\% & 100.00\% \\
MVTec AD & 0.05 & Upper quantile & Fixed SQ 25\% & 39 & 39 & 39 & 0 & 0 & 100.00\% & 100.00\% \\
VisA & 0.01 & Median & Fixed SQ 0.5\% & 23 & 36 & 36 & 0 & 0 & 63.89\% & 63.89\% \\
VisA & 0.01 & Upper quantile & Fixed SQ 0.5\% & 36 & 36 & 36 & 0 & 0 & 100.00\% & 100.00\% \\
VisA & 0.01 & Median & Fixed SQ 1\% & 26 & 36 & 36 & 0 & 0 & 72.22\% & 72.22\% \\
VisA & 0.01 & Upper quantile & Fixed SQ 1\% & 36 & 36 & 36 & 0 & 0 & 100.00\% & 100.00\% \\
VisA & 0.01 & Median & Fixed SQ 5\% & 31 & 36 & 36 & 0 & 0 & 86.11\% & 86.11\% \\
VisA & 0.01 & Upper quantile & Fixed SQ 5\% & 36 & 36 & 36 & 0 & 0 & 100.00\% & 100.00\% \\
VisA & 0.01 & Median & Fixed SQ 10\% & 33 & 36 & 36 & 0 & 0 & 91.67\% & 91.67\% \\
VisA & 0.01 & Upper quantile & Fixed SQ 10\% & 36 & 36 & 36 & 0 & 0 & 100.00\% & 100.00\% \\
VisA & 0.01 & Median & Fixed SQ 25\% & 36 & 36 & 36 & 0 & 0 & 100.00\% & 100.00\% \\
VisA & 0.01 & Upper quantile & Fixed SQ 25\% & 36 & 36 & 36 & 0 & 0 & 100.00\% & 100.00\% \\
VisA & 0.05 & Median & Fixed SQ 0.5\% & 13 & 36 & 36 & 0 & 0 & 36.11\% & 36.11\% \\
VisA & 0.05 & Upper quantile & Fixed SQ 0.5\% & 34 & 36 & 36 & 0 & 0 & 94.44\% & 94.44\% \\
VisA & 0.05 & Median & Fixed SQ 1\% & 19 & 36 & 36 & 0 & 0 & 52.78\% & 52.78\% \\
VisA & 0.05 & Upper quantile & Fixed SQ 1\% & 35 & 36 & 36 & 0 & 0 & 97.22\% & 97.22\% \\
VisA & 0.05 & Median & Fixed SQ 5\% & 29 & 36 & 36 & 0 & 0 & 80.56\% & 80.56\% \\
VisA & 0.05 & Upper quantile & Fixed SQ 5\% & 36 & 36 & 36 & 0 & 0 & 100.00\% & 100.00\% \\
VisA & 0.05 & Median & Fixed SQ 10\% & 30 & 36 & 36 & 0 & 0 & 83.33\% & 83.33\% \\
VisA & 0.05 & Upper quantile & Fixed SQ 10\% & 36 & 36 & 36 & 0 & 0 & 100.00\% & 100.00\% \\
VisA & 0.05 & Median & Fixed SQ 25\% & 36 & 36 & 36 & 0 & 0 & 100.00\% & 100.00\% \\
VisA & 0.05 & Upper quantile & Fixed SQ 25\% & 36 & 36 & 36 & 0 & 0 & 100.00\% & 100.00\% \\

\end{longtable}

\endgroup

\end{landscape}

\clearpage

\begin{landscape}

\section{Original-bank backbone results}

\begingroup

\footnotesize

\setlength{\tabcolsep}{3pt}

\renewcommand{\arraystretch}{1.15}

\setcounter{table}{3}

\begin{longtable}{>{\raggedright\arraybackslash}p{\dimexpr 0.140000\linewidth-4.200000pt\relax}>{\raggedright\arraybackslash}p{\dimexpr 0.075000\linewidth-2.250000pt\relax}>{\raggedright\arraybackslash}p{\dimexpr 0.365000\linewidth-10.950000pt\relax}>{\raggedright\arraybackslash}p{\dimexpr 0.210000\linewidth-6.300000pt\relax}>{\raggedright\arraybackslash}p{\dimexpr 0.210000\linewidth-6.300000pt\relax}}

\caption{Original-bank backbone comparison for decision disagreement, calibration attainment, ranking and paired tests.}\label{tab:s4}\\

\toprule

Dataset & \(\alpha\) & Metric & WideResNet-50 & ResNet-18 \\

\midrule

\endfirsthead

\toprule

Dataset & \(\alpha\) & Metric & WideResNet-50 & ResNet-18 \\

\midrule

\endhead

\bottomrule

\endfoot

VisA & 0.01 & Dcomp \(\cdot\) Prefix & 0.01022 & 0.00911 \\

VisA & 0.01 & Dcomp \(\cdot\) Random & 0.01970 & 0.02093 \\

VisA & 0.01 & Dcomp \(\cdot\) Rebuilt greedy & 0.00963 & 0.00891 \\

VisA & 0.01 & Dcomp \(\cdot\) Budget-matched PQ & 0.00298 & 0.00255 \\

VisA & 0.01 & Median attainment & 35/36 & 35/36 \\

VisA & 0.01 & Upper-quantile attainment & 36/36 & 36/36 \\

VisA & 0.01 & Category-averaged I-AUROC loss & 0.000276 & 0.003177 \\

VisA & 0.01 & Anomaly nearest-bin flips & 81/288 (28.12\%) & 76/288 (26.39\%) \\

VisA & 0.01 & Normal holdout flips & 15/1152 (1.30\%) & 10/1152 (0.87\%) \\

VisA & 0.01 & Holm p \(\cdot\) PQ vs Prefix & 0.0244 & 0.00146 \\

VisA & 0.01 & Holm p \(\cdot\) PQ vs Random & 0.0244 & 0.00146 \\

VisA & 0.01 & Holm p \(\cdot\) PQ vs Rebuilt greedy & 0.0244 & 0.00146 \\

VisA & 0.05 & Dcomp \(\cdot\) Prefix & 0.01960 & 0.02080 \\

VisA & 0.05 & Dcomp \(\cdot\) Random & 0.04028 & 0.04424 \\

VisA & 0.05 & Dcomp \(\cdot\) Rebuilt greedy & 0.02207 & 0.01992 \\

VisA & 0.05 & Dcomp \(\cdot\) Budget-matched PQ & 0.00529 & 0.00623 \\

VisA & 0.05 & Median attainment & 35/36 & 35/36 \\

VisA & 0.05 & Upper-quantile attainment & 36/36 & 36/36 \\

VisA & 0.05 & Category-averaged I-AUROC loss & 0.000276 & 0.003177 \\

VisA & 0.05 & Anomaly nearest-bin flips & 96/288 (33.33\%) & 99/288 (34.38\%) \\

VisA & 0.05 & Normal holdout flips & 41/1152 (3.56\%) & 26/1152 (2.26\%) \\

VisA & 0.05 & Holm p \(\cdot\) PQ vs Prefix & 0.0244 & 0.00586 \\

VisA & 0.05 & Holm p \(\cdot\) PQ vs Random & 0.0244 & 0.00146 \\

VisA & 0.05 & Holm p \(\cdot\) PQ vs Rebuilt greedy & 0.0244 & 0.00586 \\

MVTec AD & 0.01 & Dcomp \(\cdot\) Prefix & 0.06988 & 0.05683 \\

MVTec AD & 0.01 & Dcomp \(\cdot\) Random & 0.16876 & 0.22323 \\

MVTec AD & 0.01 & Dcomp \(\cdot\) Rebuilt greedy & 0.06477 & 0.06057 \\

MVTec AD & 0.01 & Dcomp \(\cdot\) Budget-matched PQ & 0.01400 & 0.01477 \\

MVTec AD & 0.01 & Median attainment & 34/39 & 34/39 \\

MVTec AD & 0.01 & Upper-quantile attainment & 39/39 & 39/39 \\

MVTec AD & 0.01 & Category-averaged I-AUROC loss & 0.001513 & 0.003335 \\

MVTec AD & 0.01 & Anomaly nearest-bin flips & 253/654 (38.68\%) & 279/654 (42.66\%) \\

MVTec AD & 0.01 & Normal holdout flips & 44/1248 (3.53\%) & 42/1248 (3.37\%) \\

MVTec AD & 0.01 & Holm p \(\cdot\) PQ vs Prefix & 0.0200 & 0.000732 \\

MVTec AD & 0.01 & Holm p \(\cdot\) PQ vs Random & 0.0122 & 0.000732 \\

MVTec AD & 0.01 & Holm p \(\cdot\) PQ vs Rebuilt greedy & 0.0200 & 0.000732 \\

MVTec AD & 0.05 & Dcomp \(\cdot\) Prefix & 0.06429 & 0.08452 \\

MVTec AD & 0.05 & Dcomp \(\cdot\) Random & 0.17224 & 0.26342 \\

MVTec AD & 0.05 & Dcomp \(\cdot\) Rebuilt greedy & 0.06217 & 0.08282 \\

MVTec AD & 0.05 & Dcomp \(\cdot\) Budget-matched PQ & 0.01193 & 0.02375 \\

MVTec AD & 0.05 & Median attainment & 32/39 & 33/39 \\

MVTec AD & 0.05 & Upper-quantile attainment & 38/39 & 39/39 \\

MVTec AD & 0.05 & Category-averaged I-AUROC loss & 0.001513 & 0.003335 \\

MVTec AD & 0.05 & Anomaly nearest-bin flips & 247/654 (37.77\%) & 301/654 (46.02\%) \\

MVTec AD & 0.05 & Normal holdout flips & 67/1248 (5.37\%) & 104/1248 (8.33\%) \\

MVTec AD & 0.05 & Holm p \(\cdot\) PQ vs Prefix & 0.0122 & 0.000732 \\

MVTec AD & 0.05 & Holm p \(\cdot\) PQ vs Random & 0.0122 & 0.000732 \\

MVTec AD & 0.05 & Holm p \(\cdot\) PQ vs Rebuilt greedy & 0.0190 & 0.000732 \\

\end{longtable}

\endgroup

\end{landscape}

\clearpage

\begin{landscape}

\section{Backbone and reference-bank results}

\begingroup

\footnotesize

\setlength{\tabcolsep}{3pt}

\renewcommand{\arraystretch}{1.15}

\setcounter{table}{4}

\begin{longtable}{>{\raggedright\arraybackslash}p{\dimexpr 0.430000\linewidth-10.320000pt\relax}>{\raggedright\arraybackslash}p{\dimexpr 0.190000\linewidth-4.560000pt\relax}>{\raggedright\arraybackslash}p{\dimexpr 0.190000\linewidth-4.560000pt\relax}>{\raggedright\arraybackslash}p{\dimexpr 0.190000\linewidth-4.560000pt\relax}}

\caption{Backbone and reference bank comparison. Attainment counts are shown as median and upper-quantile counts separated by a centered dot.}\label{tab:s5}\\

\toprule

Measure & WideResNet-50 original & ResNet-18 original & ResNet-18 clean \\

\midrule

\endfirsthead

\toprule

Measure & WideResNet-50 original & ResNet-18 original & ResNet-18 clean \\

\midrule

\endhead

\bottomrule

\endfoot

J1 minimum Dcomp & 4/4 & 4/4 & 4/4 \\

J2 joint calibration attainment & 4/4 & 4/4 & 4/4 \\

J3 nearest-bin anomaly > holdout & 4/4 & 4/4 & 4/4 \\

VisA, \(\alpha\) = 1\%, median / upper quantile & 35/36 \(\cdot\) 36/36 & 35/36 \(\cdot\) 36/36 & 35/36 \(\cdot\) 36/36 \\

VisA, \(\alpha\) = 5\%, median / upper quantile & 35/36 \(\cdot\) 36/36 & 35/36 \(\cdot\) 36/36 & 36/36 \(\cdot\) 36/36 \\

MVTec AD, \(\alpha\) = 1\%, median / upper quantile & 34/39 \(\cdot\) 39/39 & 34/39 \(\cdot\) 39/39 & 32/39 \(\cdot\) 39/39 \\

MVTec AD, \(\alpha\) = 5\%, median / upper quantile & 32/39 \(\cdot\) 38/39 & 33/39 \(\cdot\) 39/39 & 36/39 \(\cdot\) 39/39 \\

VisA full-bank I-AUROC & 0.911357 & 0.910232 & 0.915523 \\

VisA PQ loss & 0.000276 & 0.003177 & 0.003382 \\

MVTec AD full-bank I-AUROC & 0.977403 & 0.946652 & 0.953251 \\

MVTec AD PQ loss & 0.001513 & 0.003335 & 0.003715 \\

\end{longtable}

\endgroup

\end{landscape}

\clearpage

\begin{landscape}

\section{Deployment size and ranking measurements}

\begingroup

\footnotesize

\setlength{\tabcolsep}{3pt}

\renewcommand{\arraystretch}{1.15}

\setcounter{table}{5}

\begin{longtable}{>{\raggedright\arraybackslash}p{\dimexpr 0.140000\linewidth-5.040000pt\relax}>{\raggedright\arraybackslash}p{\dimexpr 0.250000\linewidth-9.000000pt\relax}>{\raggedright\arraybackslash}p{\dimexpr 0.180000\linewidth-6.480000pt\relax}>{\raggedright\arraybackslash}p{\dimexpr 0.170000\linewidth-6.120000pt\relax}>{\raggedright\arraybackslash}p{\dimexpr 0.140000\linewidth-5.040000pt\relax}>{\raggedright\arraybackslash}p{\dimexpr 0.120000\linewidth-4.320000pt\relax}}

\caption{Realized deployment size and image-level AUROC on the original WideResNet-50 banks.}\label{tab:s6}\\

\toprule

Dataset & Method & Nominal fraction & Mean bytes & I-AUROC & Categories \\

\midrule

\endfirsthead

\toprule

Dataset & Method & Nominal fraction & Mean bytes & I-AUROC & Categories \\

\midrule

\endhead

\bottomrule

\endfoot

VisA & Prefix & 0.005 & 106520 & 0.601200 & 12 \\

VisA & Prefix & 0.01 & 217112 & 0.729600 & 12 \\

VisA & Prefix & 0.05 & 1089560 & 0.877600 & 12 \\

VisA & Prefix & 0.10 & 2183192 & 0.900000 & 12 \\

VisA & Prefix & 0.25 & 5455896 & 0.908800 & 12 \\

VisA & Random & 0.005 & 106520 & 0.556400 & 12 \\

VisA & Random & 0.01 & 217112 & 0.574300 & 12 \\

VisA & Random & 0.05 & 1089560 & 0.752100 & 12 \\

VisA & Random & 0.10 & 2183192 & 0.814000 & 12 \\

VisA & Random & 0.25 & 5455896 & 0.878700 & 12 \\

VisA & Rebuilt greedy & 0.005 & 106520 & 0.662200 & 12 \\

VisA & Rebuilt greedy & 0.01 & 217112 & 0.741500 & 12 \\

VisA & Rebuilt greedy & 0.05 & 1089560 & 0.873700 & 12 \\

VisA & Rebuilt greedy & 0.10 & 2183192 & 0.891300 & 12 \\

VisA & Rebuilt greedy & 0.25 & 5455896 & 0.908100 & 12 \\

VisA & Budget-matched PQ & 0.005 & 108294 & 0.758900 & 12 \\

VisA & Budget-matched PQ & 0.01 & 217474 & 0.840400 & 12 \\

VisA & Budget-matched PQ & 0.05 & 1091334 & 0.882100 & 12 \\

VisA & Budget-matched PQ & 0.10 & 2180360 & 0.911100 & 12 \\

VisA & Budget-matched PQ & 0.25 & 5453191 & 0.911300 & 12 \\

VisA & Int8 SQ & 0.005 & 108568 & 0.812601 & 12 \\

VisA & Int8 SQ & 0.01 & 218136 & 0.871654 & 12 \\

VisA & Int8 SQ & 0.05 & 1091608 & 0.904728 & 12 \\

VisA & Int8 SQ & 0.10 & 2183192 & 0.911696 & 12 \\

VisA & Int8 SQ & 0.25 & 5457944 & 0.911296 & 12 \\

MVTec AD & Prefix & 0.005 & 106520 & 0.713200 & 13 \\

MVTec AD & Prefix & 0.01 & 217112 & 0.844900 & 13 \\

MVTec AD & Prefix & 0.05 & 1089560 & 0.949100 & 13 \\

MVTec AD & Prefix & 0.10 & 2183192 & 0.967300 & 13 \\

MVTec AD & Prefix & 0.25 & 5455896 & 0.975500 & 13 \\

MVTec AD & Random & 0.005 & 106520 & 0.704900 & 13 \\

MVTec AD & Random & 0.01 & 217112 & 0.744700 & 13 \\

MVTec AD & Random & 0.05 & 1089560 & 0.830000 & 13 \\

MVTec AD & Random & 0.10 & 2183192 & 0.880800 & 13 \\

MVTec AD & Random & 0.25 & 5455896 & 0.946300 & 13 \\

MVTec AD & Rebuilt greedy & 0.005 & 106520 & 0.778100 & 13 \\

MVTec AD & Rebuilt greedy & 0.01 & 217112 & 0.857400 & 13 \\

MVTec AD & Rebuilt greedy & 0.05 & 1089560 & 0.953100 & 13 \\

MVTec AD & Rebuilt greedy & 0.10 & 2183192 & 0.970400 & 13 \\

MVTec AD & Rebuilt greedy & 0.25 & 5455896 & 0.975400 & 13 \\

MVTec AD & Budget-matched PQ & 0.005 & 108294 & 0.818900 & 13 \\

MVTec AD & Budget-matched PQ & 0.01 & 217474 & 0.898900 & 13 \\

MVTec AD & Budget-matched PQ & 0.05 & 1091334 & 0.943400 & 13 \\

MVTec AD & Budget-matched PQ & 0.10 & 2180360 & 0.975900 & 13 \\

MVTec AD & Budget-matched PQ & 0.25 & 5453191 & 0.977100 & 13 \\

MVTec AD & Int8 SQ & 0.005 & 108568 & 0.904289 & 13 \\

MVTec AD & Int8 SQ & 0.01 & 218136 & 0.936851 & 13 \\

MVTec AD & Int8 SQ & 0.05 & 1091608 & 0.974676 & 13 \\

MVTec AD & Int8 SQ & 0.10 & 2183192 & 0.976944 & 13 \\

MVTec AD & Int8 SQ & 0.25 & 5457944 & 0.977425 & 13 \\

VisA & Lightweight PQ & actual single point & 108208 & 0.721300 & 12 \\

MVTec AD & Lightweight PQ & actual single point & 108208 & 0.802500 & 13 \\

VisA & Full bank & 1.0 & 21835800 & 0.911400 & 12 \\

MVTec AD & Full bank & 1.0 & 21835800 & 0.977400 & 13 \\

\end{longtable}

\endgroup

\end{landscape}

\clearpage

\begin{landscape}

\section{Compression representations and parameters}

\begingroup

\footnotesize

\setlength{\tabcolsep}{3pt}

\renewcommand{\arraystretch}{1.15}

\setcounter{table}{6}

\begin{longtable}{>{\raggedright\arraybackslash}p{\dimexpr 0.144231\linewidth-5.192308pt\relax}>{\raggedright\arraybackslash}p{\dimexpr 0.201923\linewidth-7.269231pt\relax}>{\raggedright\arraybackslash}p{\dimexpr 0.192308\linewidth-6.923077pt\relax}>{\raggedright\arraybackslash}p{\dimexpr 0.259615\linewidth-9.346154pt\relax}>{\raggedright\arraybackslash}p{\dimexpr 0.100962\linewidth-3.634615pt\relax}>{\raggedright\arraybackslash}p{\dimexpr 0.100962\linewidth-3.634615pt\relax}}

\caption{Compression methods compared at matched deployment sizes. Byte columns give deployed sizes at the 10\% tier; the full reference bank and lightweight PQ are listed at their single measured sizes.}\label{tab:s7}\\

\toprule

Method & Stored representation & Budget allocation & Key parameters & VisA bytes & MVTec AD bytes \\

\midrule

\endfirsthead

\toprule

Method & Stored representation & Budget allocation & Key parameters & VisA bytes & MVTec AD bytes \\

\midrule

\endhead

\bottomrule

\endfoot

Full reference bank & All rows as float32 vectors & Uncapped; defines \(B_{0}\) & 5331 rows \(\times\) 1024 dimensions; 24-byte header & 21835800 & 21835800 \\

Prefix & Leading rows in bank order, float32 & Rows that fit under the cap after the header & 533 rows at the 10\% tier & 2183192 & 2183192 \\

Random & Rows drawn without replacement, stored in bank order, float32 & Rows that fit under the cap after the header & 533 rows at the 10\% tier; sampling seed 17 & 2183192 & 2183192 \\

Rebuilt greedy & Rows added by greatest squared distance to the selected set, float32 & Rows that fit under the cap after the header & 533 rows at the 10\% tier; selection starts from the first row & 2183192 & 2183192 \\

Budget-matched PQ & PQ codes for every row with per-subspace codebooks & Layout chosen under the cap, code precision first, then payload size & 190 subspaces of width 6 and 256 centroids per subspace at the 10\% tier; codebooks trained on 4096 rows with 8 k-means iterations & 2180360 & 2180360 \\

Int8 SQ & Leading rows as signed 8-bit codes with per-dimension float32 minimum and range & Leading rows that fit under the cap with the two per-dimension vectors & 8 bits per coordinate; 2124 rows at the 10\% tier & 2183192 & 2183192 \\

Lightweight PQ & PQ codes for every row with a fixed codebook layout & Fixed size, independent of the tier & 8 subspaces and 16 centroids per subspace & 108208 & 108208 \\

\end{longtable}

\endgroup

\end{landscape}

\clearpage

\begin{landscape}

\section{Dataset coverage and evaluation roles}

\begingroup

\footnotesize

\setlength{\tabcolsep}{3pt}

\renewcommand{\arraystretch}{1.15}

\setcounter{table}{7}

\begin{longtable}{>{\raggedright\arraybackslash}p{\dimexpr 0.095000\linewidth-6.270000pt\relax}>{\raggedright\arraybackslash}p{\dimexpr 0.100000\linewidth-6.600000pt\relax}>{\raggedright\arraybackslash}p{\dimexpr 0.065000\linewidth-4.290000pt\relax}>{\raggedright\arraybackslash}p{\dimexpr 0.050000\linewidth-3.300000pt\relax}>{\raggedright\arraybackslash}p{\dimexpr 0.125000\linewidth-8.250000pt\relax}>{\raggedright\arraybackslash}p{\dimexpr 0.085000\linewidth-5.610000pt\relax}>{\raggedright\arraybackslash}p{\dimexpr 0.105000\linewidth-6.930000pt\relax}>{\raggedright\arraybackslash}p{\dimexpr 0.090000\linewidth-5.940000pt\relax}>{\raggedright\arraybackslash}p{\dimexpr 0.085000\linewidth-5.610000pt\relax}>{\raggedright\arraybackslash}p{\dimexpr 0.105000\linewidth-6.930000pt\relax}>{\raggedright\arraybackslash}p{\dimexpr 0.095000\linewidth-6.270000pt\relax}}

\caption{Dataset coverage and calibration and holdout roles. Em dashes mark independent-pool summaries, which are available for VisA and MVTec AD; MPDD has no records in that pool inventory.}\label{tab:s8}\\

\toprule

Dataset & Categories & Seeds & Cells & Independent-pool records & Eligible normal images (total) & Calibration pool (total) & Normal holdout (total) & Eligible normal images (mean) & Calibration pool (mean) & Normal holdout (mean) \\

\midrule

\endfirsthead

\toprule

Dataset & Categories & Seeds & Cells & Independent-pool records & Eligible normal images (total) & Calibration pool (total) & Normal holdout (total) & Eligible normal images (mean) & Calibration pool (mean) & Normal holdout (mean) \\

\midrule

\endhead

\bottomrule

\endfoot

VisA & 12 & 17, 29, 43 & 36 & 36 & 14117 & 12965 & 1152 & 392.1389 & 360.1389 & 32.0000 \\

MVTec AD & 13 & 17, 29, 43 & 39 & 39 & 6201 & 4953 & 1248 & 159.0000 & 127.0000 & 32.0000 \\

MPDD & 6 & 17, 29, 43 & 18 & 0 & — & — & — & — & — & — \\

\end{longtable}

\endgroup

\end{landscape}

\clearpage

\begin{landscape}

\section{Experimental settings}

\begingroup

\footnotesize

\setlength{\tabcolsep}{3pt}

\renewcommand{\arraystretch}{1.15}

\setcounter{table}{8}

\begin{longtable}{>{\raggedright\arraybackslash}p{\dimexpr 0.280000\linewidth-3.360000pt\relax}>{\raggedright\arraybackslash}p{\dimexpr 0.720000\linewidth-8.640000pt\relax}}

\caption{Experimental settings of the primary comparison and the validation comparisons.}\label{tab:s9}\\

\toprule

Item & Setting \\

\midrule

\endfirsthead

\toprule

Item & Setting \\

\midrule

\endhead

\bottomrule

\endfoot

Datasets and cells & VisA: 12 categories, 36 cells; MVTec AD: 13 categories, 39 cells; MPDD: 6 categories, 18 cells \\

Seeds & 17, 29 and 43 for every category \\

Primary backbone and feature layers & ImageNet-pretrained WideResNet-50; second and third residual groups, 28 \(\times\) 28 patch grid, 1024-dimensional descriptors \\

Second backbone & ImageNet-pretrained ResNet-18 on VisA and MVTec AD with the original and clean banks at the 1\%, 5\% and 10\% tiers \\

Input size & Shorter edge resized to 256 pixels, 224 \(\times\) 224 center crop, ImageNet normalization \\

Budget tiers & 0.5\%, 1\%, 5\%, 10\% and 25\% of the full-bank vector payload \\

Operating levels & \(\alpha\) = 1\% and 5\% \\

Recalibrations & 200 draws of 32 images from the calibration pool P in each cell \\

Probe set & 32 images per cell; thresholds at ranks 32 and 31 for \(\alpha\) = 1\% and 5\% \\

Calibration pool and normal holdout & VisA: 12965 and 1152 images (360.1 and 32.0 per cell); MVTec AD: 4953 and 1248 images (127.0 and 32.0 per cell) \\

Intervals & Category bootstrap with 10000 resamples for category-equally weighted means, attainment fractions and paired differences; Wilson score interval for pooled proportions \\

Tests & Exact sign-flip test on paired category differences with Holm adjustment within each test family \\

\end{longtable}

\endgroup

\end{landscape}

\clearpage

\begin{landscape}

\section{Fixed-budget and automatic-selection attainment}

\begingroup

\footnotesize

\setlength{\tabcolsep}{3pt}

\renewcommand{\arraystretch}{1.00}

\setcounter{table}{9}

\begin{longtable}{>{\raggedright\arraybackslash}p{\dimexpr 0.085000\linewidth-5.610000pt\relax}>{\raggedright\arraybackslash}p{\dimexpr 0.065000\linewidth-4.290000pt\relax}>{\raggedright\arraybackslash}p{\dimexpr 0.115000\linewidth-7.590000pt\relax}>{\raggedright\arraybackslash}p{\dimexpr 0.180000\linewidth-11.880000pt\relax}>{\raggedright\arraybackslash}p{\dimexpr 0.055000\linewidth-3.630000pt\relax}>{\raggedright\arraybackslash}p{\dimexpr 0.075000\linewidth-4.950000pt\relax}>{\raggedright\arraybackslash}p{\dimexpr 0.075000\linewidth-4.950000pt\relax}>{\raggedright\arraybackslash}p{\dimexpr 0.045000\linewidth-2.970000pt\relax}>{\raggedright\arraybackslash}p{\dimexpr 0.085000\linewidth-5.610000pt\relax}>{\raggedright\arraybackslash}p{\dimexpr 0.110000\linewidth-7.260000pt\relax}>{\raggedright\arraybackslash}p{\dimexpr 0.110000\linewidth-7.260000pt\relax}}

\caption{Calibration benchmark attainment under fixed budgets and automatic selection, including int8 SQ at the 10\% and 25\% tiers.}\label{tab:s10}\\

\toprule

Dataset & \(\alpha\) & Benchmark & Configuration & Pass & Eligible & Planned & NA & Fallback & Eligible rate & Planned rate \\

\midrule

\endfirsthead

\toprule

Dataset & \(\alpha\) & Benchmark & Configuration & Pass & Eligible & Planned & NA & Fallback & Eligible rate & Planned rate \\

\midrule

\endhead

\bottomrule

\endfoot

MPDD & 0.01 & Median & Fixed PQ 1\% & 1 & 12 & 18 & 6 & 0 & 8.33\% & 5.56\% \\

MPDD & 0.01 & Upper quantile & Fixed PQ 1\% & 5 & 12 & 18 & 6 & 0 & 41.67\% & 27.78\% \\

MPDD & 0.01 & Median & Fixed PQ 10\% & 9 & 15 & 18 & 3 & 0 & 60.00\% & 50.00\% \\

MPDD & 0.01 & Upper quantile & Fixed PQ 10\% & 10 & 15 & 18 & 3 & 0 & 66.67\% & 55.56\% \\

MPDD & 0.01 & Median & Automatic selection PQ & 8 & 15 & 18 & 3 & 2 & 53.33\% & 44.44\% \\

MPDD & 0.01 & Upper quantile & Automatic selection PQ & 12 & 15 & 18 & 3 & 2 & 80.00\% & 66.67\% \\

MPDD & 0.01 & Median & Fixed SQ 10\% & 8 & 15 & 18 & 3 & 0 & 53.33\% & 44.44\% \\

MPDD & 0.01 & Upper quantile & Fixed SQ 10\% & 11 & 15 & 18 & 3 & 0 & 73.33\% & 61.11\% \\

MPDD & 0.01 & Median & Fixed SQ 25\% & 15 & 15 & 18 & 3 & 0 & 100.00\% & 83.33\% \\

MPDD & 0.01 & Upper quantile & Fixed SQ 25\% & 15 & 15 & 18 & 3 & 0 & 100.00\% & 83.33\% \\

MPDD & 0.05 & Median & Fixed PQ 1\% & 0 & 12 & 18 & 6 & 0 & 0.00\% & 0.00\% \\

MPDD & 0.05 & Upper quantile & Fixed PQ 1\% & 2 & 12 & 18 & 6 & 0 & 16.67\% & 11.11\% \\

MPDD & 0.05 & Median & Fixed PQ 10\% & 7 & 15 & 18 & 3 & 0 & 46.67\% & 38.89\% \\

MPDD & 0.05 & Upper quantile & Fixed PQ 10\% & 9 & 15 & 18 & 3 & 0 & 60.00\% & 50.00\% \\

MPDD & 0.05 & Median & Automatic selection PQ & 5 & 15 & 18 & 3 & 1 & 33.33\% & 27.78\% \\

MPDD & 0.05 & Upper quantile & Automatic selection PQ & 8 & 15 & 18 & 3 & 1 & 53.33\% & 44.44\% \\

MPDD & 0.05 & Median & Fixed SQ 10\% & 4 & 15 & 18 & 3 & 0 & 26.67\% & 22.22\% \\

MPDD & 0.05 & Upper quantile & Fixed SQ 10\% & 8 & 15 & 18 & 3 & 0 & 53.33\% & 44.44\% \\

MPDD & 0.05 & Median & Fixed SQ 25\% & 15 & 15 & 18 & 3 & 0 & 100.00\% & 83.33\% \\

MPDD & 0.05 & Upper quantile & Fixed SQ 25\% & 15 & 15 & 18 & 3 & 0 & 100.00\% & 83.33\% \\

MVTec AD & 0.01 & Median & Fixed PQ 1\% & 13 & 39 & 39 & 0 & 0 & 33.33\% & 33.33\% \\

MVTec AD & 0.01 & Upper quantile & Fixed PQ 1\% & 25 & 39 & 39 & 0 & 0 & 64.10\% & 64.10\% \\

MVTec AD & 0.01 & Median & Fixed PQ 10\% & 34 & 39 & 39 & 0 & 0 & 87.18\% & 87.18\% \\

MVTec AD & 0.01 & Upper quantile & Fixed PQ 10\% & 39 & 39 & 39 & 0 & 0 & 100.00\% & 100.00\% \\

MVTec AD & 0.01 & Median & Automatic selection PQ & 20 & 39 & 39 & 0 & 0 & 51.28\% & 51.28\% \\

MVTec AD & 0.01 & Upper quantile & Automatic selection PQ & 29 & 39 & 39 & 0 & 0 & 74.36\% & 74.36\% \\

MVTec AD & 0.01 & Median & Fixed SQ 10\% & 32 & 39 & 39 & 0 & 0 & 82.05\% & 82.05\% \\

MVTec AD & 0.01 & Upper quantile & Fixed SQ 10\% & 38 & 39 & 39 & 0 & 0 & 97.44\% & 97.44\% \\

MVTec AD & 0.01 & Median & Fixed SQ 25\% & 39 & 39 & 39 & 0 & 0 & 100.00\% & 100.00\% \\

MVTec AD & 0.01 & Upper quantile & Fixed SQ 25\% & 39 & 39 & 39 & 0 & 0 & 100.00\% & 100.00\% \\

MVTec AD & 0.05 & Median & Fixed PQ 1\% & 6 & 39 & 39 & 0 & 0 & 15.38\% & 15.38\% \\

MVTec AD & 0.05 & Upper quantile & Fixed PQ 1\% & 15 & 39 & 39 & 0 & 0 & 38.46\% & 38.46\% \\

MVTec AD & 0.05 & Median & Fixed PQ 10\% & 32 & 39 & 39 & 0 & 0 & 82.05\% & 82.05\% \\

MVTec AD & 0.05 & Upper quantile & Fixed PQ 10\% & 38 & 39 & 39 & 0 & 0 & 97.44\% & 97.44\% \\

MVTec AD & 0.05 & Median & Automatic selection PQ & 19 & 39 & 39 & 0 & 1 & 48.72\% & 48.72\% \\

MVTec AD & 0.05 & Upper quantile & Automatic selection PQ & 29 & 39 & 39 & 0 & 1 & 74.36\% & 74.36\% \\

MVTec AD & 0.05 & Median & Fixed SQ 10\% & 21 & 39 & 39 & 0 & 0 & 53.85\% & 53.85\% \\

MVTec AD & 0.05 & Upper quantile & Fixed SQ 10\% & 39 & 39 & 39 & 0 & 0 & 100.00\% & 100.00\% \\

MVTec AD & 0.05 & Median & Fixed SQ 25\% & 39 & 39 & 39 & 0 & 0 & 100.00\% & 100.00\% \\

MVTec AD & 0.05 & Upper quantile & Fixed SQ 25\% & 39 & 39 & 39 & 0 & 0 & 100.00\% & 100.00\% \\

VisA & 0.01 & Median & Fixed PQ 1\% & 22 & 36 & 36 & 0 & 0 & 61.11\% & 61.11\% \\

VisA & 0.01 & Upper quantile & Fixed PQ 1\% & 35 & 36 & 36 & 0 & 0 & 97.22\% & 97.22\% \\

VisA & 0.01 & Median & Fixed PQ 10\% & 35 & 36 & 36 & 0 & 0 & 97.22\% & 97.22\% \\

VisA & 0.01 & Upper quantile & Fixed PQ 10\% & 36 & 36 & 36 & 0 & 0 & 100.00\% & 100.00\% \\

VisA & 0.01 & Median & Automatic selection PQ & 23 & 36 & 36 & 0 & 0 & 63.89\% & 63.89\% \\

VisA & 0.01 & Upper quantile & Automatic selection PQ & 36 & 36 & 36 & 0 & 0 & 100.00\% & 100.00\% \\

VisA & 0.01 & Median & Fixed SQ 10\% & 33 & 36 & 36 & 0 & 0 & 91.67\% & 91.67\% \\

VisA & 0.01 & Upper quantile & Fixed SQ 10\% & 36 & 36 & 36 & 0 & 0 & 100.00\% & 100.00\% \\

VisA & 0.01 & Median & Fixed SQ 25\% & 36 & 36 & 36 & 0 & 0 & 100.00\% & 100.00\% \\

VisA & 0.01 & Upper quantile & Fixed SQ 25\% & 36 & 36 & 36 & 0 & 0 & 100.00\% & 100.00\% \\

VisA & 0.05 & Median & Fixed PQ 1\% & 17 & 36 & 36 & 0 & 0 & 47.22\% & 47.22\% \\

VisA & 0.05 & Upper quantile & Fixed PQ 1\% & 33 & 36 & 36 & 0 & 0 & 91.67\% & 91.67\% \\

VisA & 0.05 & Median & Fixed PQ 10\% & 35 & 36 & 36 & 0 & 0 & 97.22\% & 97.22\% \\

VisA & 0.05 & Upper quantile & Fixed PQ 10\% & 36 & 36 & 36 & 0 & 0 & 100.00\% & 100.00\% \\

VisA & 0.05 & Median & Automatic selection PQ & 18 & 36 & 36 & 0 & 0 & 50.00\% & 50.00\% \\

VisA & 0.05 & Upper quantile & Automatic selection PQ & 35 & 36 & 36 & 0 & 0 & 97.22\% & 97.22\% \\

VisA & 0.05 & Median & Fixed SQ 10\% & 30 & 36 & 36 & 0 & 0 & 83.33\% & 83.33\% \\

VisA & 0.05 & Upper quantile & Fixed SQ 10\% & 36 & 36 & 36 & 0 & 0 & 100.00\% & 100.00\% \\

VisA & 0.05 & Median & Fixed SQ 25\% & 36 & 36 & 36 & 0 & 0 & 100.00\% & 100.00\% \\

VisA & 0.05 & Upper quantile & Fixed SQ 25\% & 36 & 36 & 36 & 0 & 0 & 100.00\% & 100.00\% \\

\end{longtable}

\endgroup

\end{landscape}

\clearpage

\begin{landscape}

\section{Paired automatic-selection comparisons}

\begingroup

\footnotesize

\setlength{\tabcolsep}{3pt}

\renewcommand{\arraystretch}{1.15}

\setcounter{table}{10}

\begin{longtable}{>{\raggedright\arraybackslash}p{\dimexpr 0.120000\linewidth-5.760000pt\relax}>{\raggedright\arraybackslash}p{\dimexpr 0.060000\linewidth-2.880000pt\relax}>{\raggedright\arraybackslash}p{\dimexpr 0.265000\linewidth-12.720000pt\relax}>{\raggedright\arraybackslash}p{\dimexpr 0.120000\linewidth-5.760000pt\relax}>{\raggedright\arraybackslash}p{\dimexpr 0.170000\linewidth-8.160000pt\relax}>{\raggedright\arraybackslash}p{\dimexpr 0.080000\linewidth-3.840000pt\relax}>{\raggedright\arraybackslash}p{\dimexpr 0.085000\linewidth-4.080000pt\relax}>{\raggedright\arraybackslash}p{\dimexpr 0.100000\linewidth-4.800000pt\relax}}

\caption{Paired comparisons of automatic selection and fixed-budget PQ.}\label{tab:s11}\\

\toprule

Dataset & \(\alpha\) & Contrast & Mean difference & 95\% CI & p & Holm p & Categories \\

\midrule

\endfirsthead

\toprule

Dataset & \(\alpha\) & Contrast & Mean difference & 95\% CI & p & Holm p & Categories \\

\midrule

\endhead

\bottomrule

\endfoot

VisA & 0.01 & Automatic selection \(-\) Fixed PQ 1\% & \(-\)0.00120 & [\(-\)0.0131, 0.0106] & 0.814 & 0.814 & 12 \\

VisA & 0.01 & Automatic selection \(-\) Fixed PQ 10\% & 0.01790 & [0.0094, 0.0264] & 0.000488 & 0.00195 & 12 \\

VisA & 0.05 & Automatic selection \(-\) Fixed PQ 1\% & \(-\)0.01040 & [\(-\)0.0274, 0.0067] & 0.210 & 0.420 & 12 \\

VisA & 0.05 & Automatic selection \(-\) Fixed PQ 10\% & 0.02690 & [0.0200, 0.0337] & 0.000488 & 0.00195 & 12 \\

MVTec AD & 0.01 & Automatic selection \(-\) Fixed PQ 1\% & \(-\)0.03840 & [\(-\)0.1008, 0.0240] & 0.210 & 0.419 & 13 \\

MVTec AD & 0.01 & Automatic selection \(-\) Fixed PQ 10\% & 0.08920 & [0.0234, 0.1549] & 0.000244 & 0.000977 & 13 \\

MVTec AD & 0.05 & Automatic selection \(-\) Fixed PQ 1\% & \(-\)0.04170 & [\(-\)0.1001, 0.0167] & 0.151 & 0.303 & 13 \\

MVTec AD & 0.05 & Automatic selection \(-\) Fixed PQ 10\% & 0.09140 & [0.0277, 0.1551] & 0.000977 & 0.00391 & 13 \\

MPDD & 0.01 & Automatic selection \(-\) Fixed PQ 1\% & \(-\)0.13070 & [\(-\)0.3620, 0.1006] & 0.250 & 0.750 & 4 \\

MPDD & 0.01 & Automatic selection \(-\) Fixed PQ 10\% & 0.05170 & [\(-\)0.0362, 0.1396] & 0.250 & 0.750 & 5 \\

MPDD & 0.05 & Automatic selection \(-\) Fixed PQ 1\% & \(-\)0.17770 & [\(-\)0.4267, 0.0713] & 0.250 & 0.563 & 4 \\

MPDD & 0.05 & Automatic selection \(-\) Fixed PQ 10\% & 0.06590 & [\(-\)0.0127, 0.1446] & 0.188 & 0.563 & 5 \\

\end{longtable}

\endgroup

\end{landscape}

\clearpage
\section{Additional interpretation of automatic selection}
Table S11 compares automatically selected PQ with fixed-budget PQ through differences within paired categories. Automatic selection uses the normal holdout to identify the smallest available budget that meets its calibration median, with the full bank serving as a fallback when the compression candidates fail that check. In the test evaluation a positive paired difference means that the automatically selected configuration changed more full-bank decisions than its fixed-budget comparator. The pairing aligns the category used for each side of the contrast, so it measures behavior on the shared evaluation coverage. The adjusted probabilities account for the multiple comparisons within the original test families.

On VisA, the paired difference in Dcomp for automatic selection minus fixed 10\% PQ was +0.0179 at \(\alpha\) = 1\% and +0.0269 at \(\alpha\) = 5\%, with Holm-adjusted p = 0.00195 at each level. On MVTec AD the corresponding differences were +0.0892 and +0.0914, with adjusted p = 0.000977 and 0.00391, respectively. On MPDD the differences were +0.0517 and +0.0659, with adjusted p = 0.750 and 0.563. In MPDD the paired intervals include zero at both levels. Table S11 retains the fixed smaller-budget contrasts as well, allowing the selected configuration to be evaluated against each of the planned fixed choices.
These paired differences concern decision agreement on the test images after the choice of deployment. They complement the byte and ranking comparisons by showing how a budget assigned individually from normal-image evidence performs against a shared budget. The calibration attainment counts in the main-article calibration-attainment figure provide the corresponding operational comparison for the shared setting, with each cell judged against its own full-bank recalibration distribution.

The cell-level points in Fig. S3 provide another view of the proxy. Under automatic selection at \(\alpha\) = 1\% on MVTec AD, the Spearman rank correlation of normal-holdout and whole-test disagreement was \(-\)0.090. For metal\_nut at seed 29, agreement on the normal holdout was complete while the compressed bank changed 71/109 decisions across the whole test set. The plot uses one point per category and seed, so the association describes how the two disagreement rates order cells. A low normal-holdout rate and a high test rate can occur in the same cell, which makes the threshold-distance breakdown informative alongside the normal-holdout measurement. It identifies where the disagreement falls among the labeled inspection images, with the image count for each rate.

Figure S2 resolves the Dcomp of budget-matched PQ by category and tier. Every category has lower disagreement at the 25\% tier than at the 0.5\% tier, and the largest values at the 1\% tier occur in pcb3 on VisA and capsule on MVTec AD.

\clearpage

\section{Image-level ranking}

\begin{figure}[htbp]

\centering

\setcounter{figure}{0}

\includegraphics[width=\linewidth,height=.72\textheight,keepaspectratio]{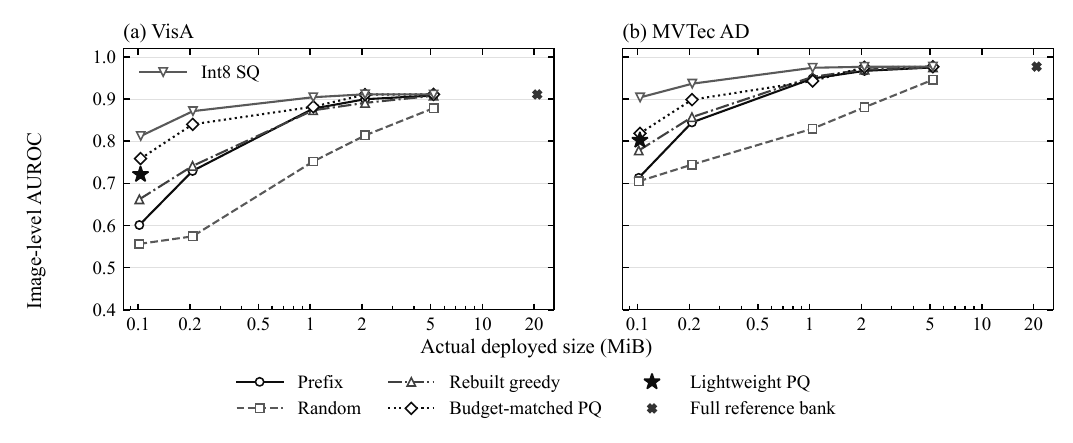}

\caption{Image-level AUROC versus deployed size for WideResNet-50. Sizes are based on the deployed representations. The full bank and lightweight PQ are shown at their respective realized sizes. Int8 scalar quantization (SQ) is plotted at the same nominal tiers.}\label{fig:s1}

\end{figure}

\clearpage

\section{Category-level disagreement}

\begin{figure}[htbp]

\centering

\setcounter{figure}{1}

\includegraphics[width=\linewidth,height=.72\textheight,keepaspectratio]{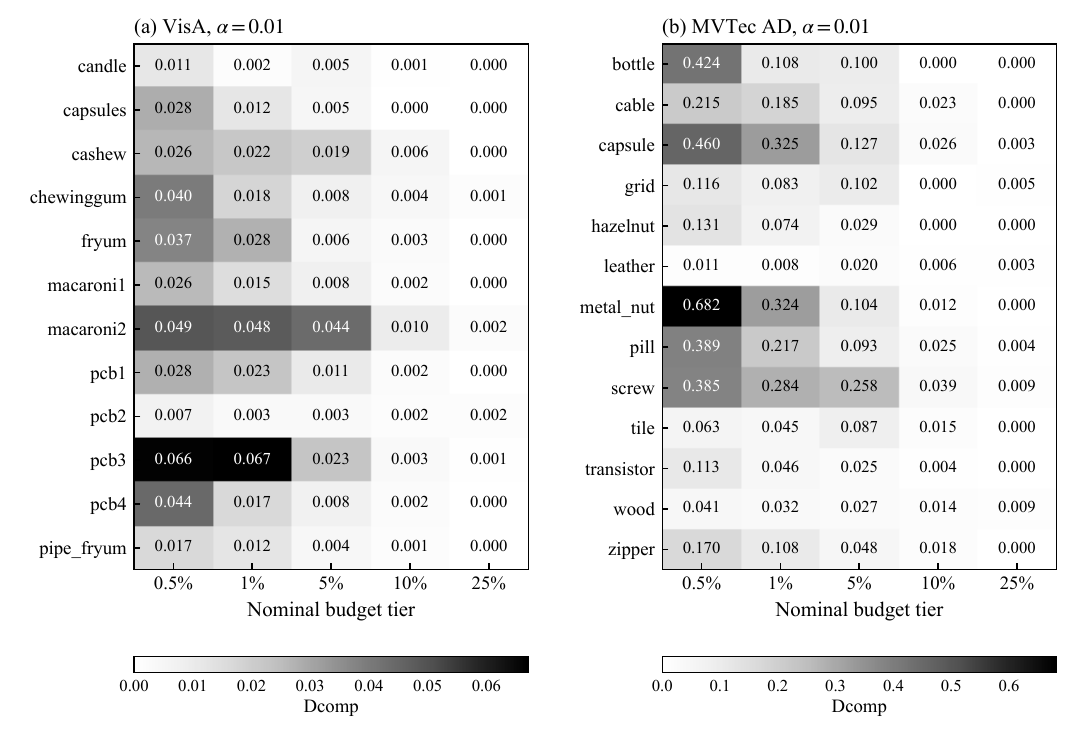}

\caption{Decision disagreement of budget-matched PQ at \(\alpha\) = 1\% by category and nominal tier for WideResNet-50. Each cell averages three seeds, and each panel uses its own gray scale.}\label{fig:s2}

\end{figure}

\clearpage

\begin{figure}[htbp]

\centering

\setcounter{figure}{2}

\includegraphics[width=\linewidth,height=.82\textheight,keepaspectratio]{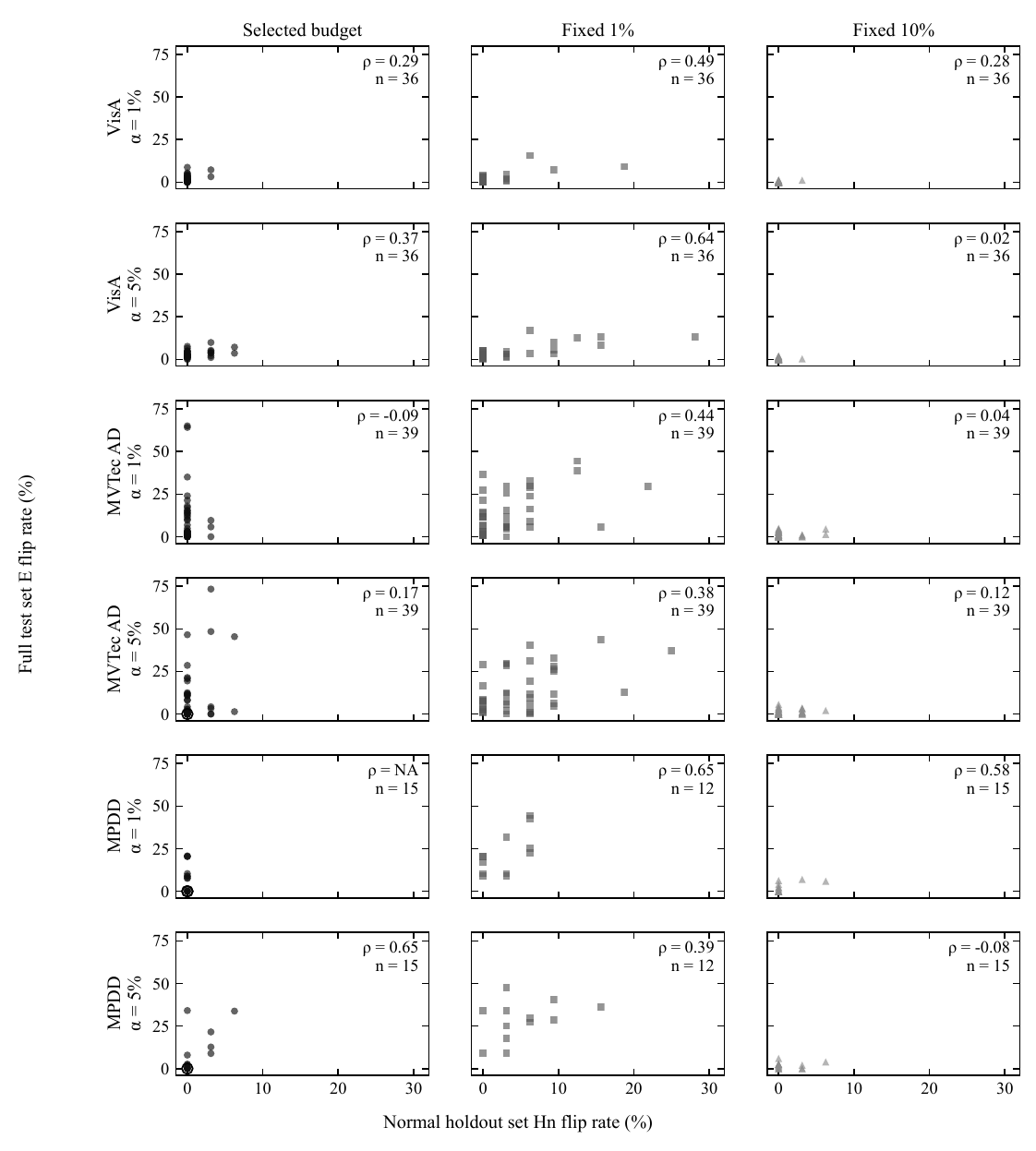}

\caption{Normal-holdout disagreement in relation to disagreement across the labeled test set in the same cells.}\label{fig:s3}

\end{figure}

\clearpage

\section{Second-backbone disagreement}

\begin{figure}[htbp]

\centering

\setcounter{figure}{3}

\includegraphics[width=\linewidth,height=.72\textheight,keepaspectratio]{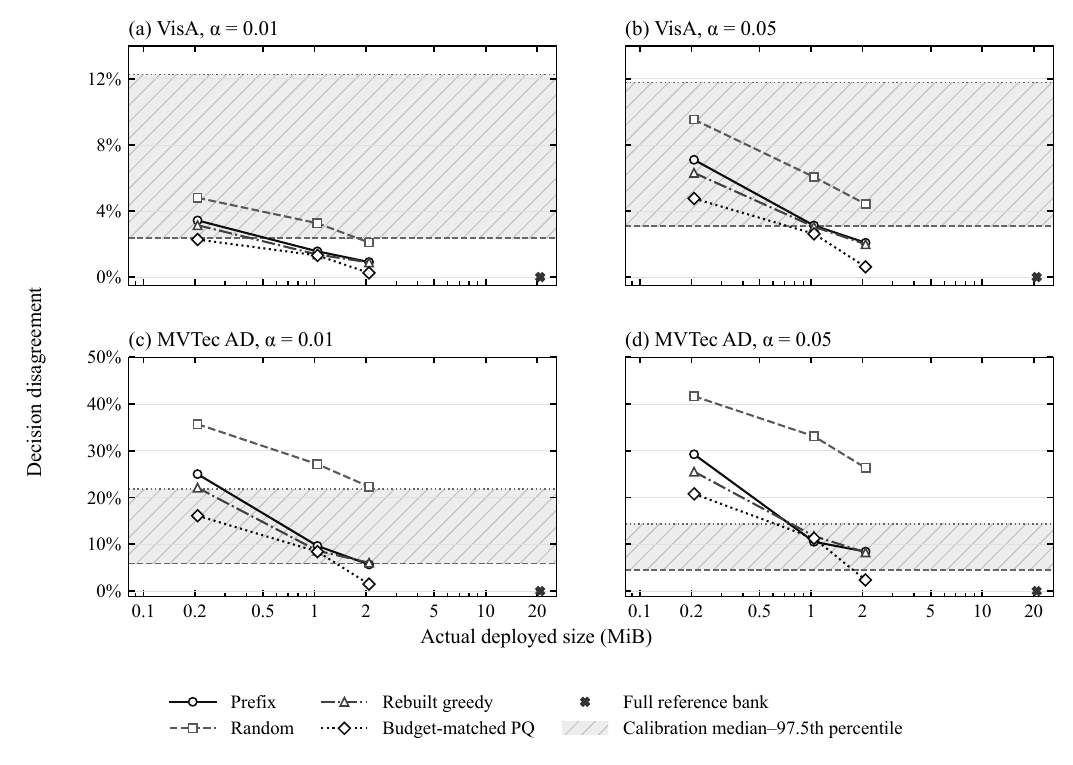}

\caption{Decision disagreement versus deployed size for ResNet-18 with the original reference banks. The VisA and MVTec AD rows use separate vertical scales.}\label{fig:s4}

\end{figure}

\end{document}